\documentclass[11pt]{article}

\usepackage[T1]{fontenc}
\usepackage{lmodern}
\usepackage[utf8]{inputenc}

\usepackage[a4paper,margin=1in]{geometry}

\usepackage{amsmath,amssymb,amsthm,mathtools}
\numberwithin{equation}{section}

\newtheorem{definition}{Definition}[section]

\usepackage{graphicx}
\usepackage{ifpdf}
\ifpdf
  \DeclareGraphicsExtensions{.pdf,.png,.jpg}
\else
  \DeclareGraphicsExtensions{.eps}
\fi
\graphicspath{{Figures/}{figures/}{./}}
\usepackage{booktabs}
\usepackage{adjustbox}
\usepackage{siunitx}
\usepackage[font=small,labelfont=bf]{caption}
\usepackage{subcaption}
\usepackage{enumitem}

\usepackage{microtype}

\usepackage{tikz}
\usetikzlibrary{arrows.meta,positioning,calc,fit}

\usepackage{algorithm}
\usepackage{algpseudocode}
\usepackage{float}
\usepackage{placeins}

\usepackage{url}
\usepackage[numbers]{natbib}
\usepackage{doi}
\usepackage{hyperref}
\hypersetup{
  colorlinks=true,
  linkcolor=blue,
  citecolor=blue,
  urlcolor=blue,
  pdftitle={False Fixed Points: Kantian Feedback, Stable Miscalibration, and Representational Compression in LLMs},
  pdfauthor={Akira Okutomi}
}
\usepackage[capitalize,nameinlink]{cleveref}

\title{False Fixed Points: Kantian Feedback, Stable Miscalibration, and Representational Compression in LLMs}
\author{Akira Okutomi\\
\small ToppyMicroServices O\"U, Tallinn, Estonia}
\date{}

\newcommand{\Hrisk}{{H_{\mathrm{Risk}}}}

\IfFileExists{commit_hash.tex}{%
}{%
}

\begin{document}
\maketitle

\begin{abstract}
High-confidence errors in large language models are often treated as fragile failures.
We study an alternative: some errors may be \emph{false fixed points}---locally stable,
internally coherent, and confidently wrong.
This separates robustness from truth-tracking.
We develop the separation through a Kantian commitment-gate framing and a minimal linear
feedback model in which stability and correctness can diverge.
Across three open-weight models, overconfident wrong items are not systematically more
locally fragile than confidently correct items under our hidden-state sensitivity probes.
Abstention-aware self-critique reduces overconfident wrong commitments by sacrificing
coverage, and C3-R, a rule-based explicit feedback gate, sharpens that tradeoff rather
than eliminating it.
These results motivate, but do not establish, high signal-to-noise (high-SNR) inertia
and representational compression as possible mechanisms for stable miscalibration.
\end{abstract}

\AtBeginEnvironment{thebibliography}{\sloppy\raggedright}

\section{Introduction}\label{sec:intro}

High-confidence hallucinations are often interpreted as brittle internal inference:
if a small perturbation can flip the answer, the error appears locally fragile.
This paper asks the complementary question: when does stability stop tracking truth?
We study the possibility that some high-confidence errors are \emph{false fixed points}:
locally stable, internally coherent, and confidently wrong.
The model does not fail because the answer is easy to dislodge.
It fails because the wrong judgment can remain stable under the perturbations we usually use to test it.
We call this observed pattern \emph{stable miscalibration}.

We use Kant to frame self-limitation: critique checks whether the conditions for
warranted judgment are present \cite{kant1781}.
In feedback terms, critique is a regulatory step that can alter commitment,
uncertainty, and correction without treating local stability as truth.
Operationally, ``Kantian feedback'' means a commitment gate: before committing,
a policy checks whether the question is answerable from available evidence,
whether the proposed answer is conceptually coherent, and whether uncertainty
justifies abstention.
The gate asks whether the model is licensed to commit at all.

We study how epistemic stability---the conditioning and robustness of inference---can
be analyzed, quantified, and tested in control systems and language models.
The central claim is conceptual but testable: local stability and truth-tracking are distinct epistemic dimensions.
Stable errors may have several mechanisms: attractor-like dynamics,
high signal-to-noise (high-SNR) representational inertia, or compression that collapses
truth-relevant distinctions into a low-effective-rank semantic region.
The experiments below establish the audit pattern and motivate these mechanisms as targets for later diagnosis.

\begin{figure}[H]
  \centering
  \begin{tikzpicture}[x=2.9cm,y=1.65cm,every node/.style={font=\scriptsize}]
    \draw[->,thick] (0,0) -- (2.18,0) node[right] {truth-tracking};
    \draw[->,thick] (0,0) -- (0,2.18) node[above] {local stability};
    \draw[gray!45] (1,0) -- (1,2);
    \draw[gray!45] (0,1) -- (2,1);
    \node[align=center,text width=2.45cm] at (0.5,0.5) {\textbf{Brittle}\\\textbf{hallucination}\\unstable wrong};
    \node[align=center,text width=2.45cm] at (1.5,0.5) {\textbf{Fragile}\\\textbf{correctness}\\unstable true};
    \node[align=center,text width=2.45cm,fill=gray!10,inner sep=3pt] at (0.5,1.5)
      {\textbf{False fixed point}\\stable wrong};
    \node[align=center,text width=2.45cm] at (1.5,1.5) {\textbf{Warranted judgment}\\stable true};
    \node[below] at (0.14,0) {low};
    \node[below] at (2.0,0) {high};
    \node[left] at (0,0.14) {low};
    \node[left] at (0,2.0) {high};
  \end{tikzpicture}
  \caption{\textbf{Local stability and truth-tracking are distinct.}
  False fixed points occupy the stable-wrong quadrant: the response is locally inert, but
  that stability does not license the judgment as true.}
  \label{fig:stability_truth_grid}
\end{figure}
\FloatBarrier

\paragraph{Contributions and claim status.}
We make three claims, with different evidential strength.
\begin{enumerate}
  \item \textbf{Feedback-stability framing.}
  Kantian self-limitation and a linear--Gaussian closed-loop model make explicit that
  stability and warranted judgment can diverge. This is a framing and motivation, not a
  mechanistic reduction of transformers.

  \item \textbf{Negative fragility result.}
  In the main hidden-state probe, overconfidently wrong (OCW) items are not systematically
  more locally fragile than confidently correct (CC) items. Self-critical prompting lowers
  overall sensitivity, but the simple ``wrong means brittle'' story is not supported in the
  tested regimes.

  \item \textbf{Audit and feedback probes.}
  The frozen audit contains C0/C1/C2 policy runs: C0 is forced judgment, C1 is cautious
  abstention, and C2 is single-shot self-critical abstention. C1/C2 reduce overconfident
  wrong commitments by lowering coverage, with utility-dependent gains. $H_{\mathrm{proxy}}$
  is a label-aware retrospective domain-triage score, not a deployment estimator. C3 is a
  post-hoc explicit feedback-loop pilot; C3-R is a stricter held-out follow-up with
  predeclared warrant blockers and a dev-selected threshold.
\end{enumerate}
These results suggest stable miscalibration as a useful audit category, but they do not
estimate $\Hrisk_{\mathrm{LLM}}$, establish representational compression, isolate a
Kant-specific causal effect, or justify broad extrapolation to larger LLMs.
When we discuss geometry, \emph{effective rank} means the entropy-based dimensionality of a
covariance spectrum, \emph{anisotropy} means directional concentration of normalized hidden
states, \emph{local tangent-rank} means the effective rank of perturbation or rewrite
displacements, and \emph{truth-separability} means how well a representation supports a
linear or contrastive true/false direction.

\medskip

Recent reliability work now separates into several nearby streams: semantic and
neighborhood-consistency measures for hallucination detection
\cite{manakul2023selfcheckgpt,farquhar2024semanticentropy,ma2025semanticenergy,
xu2026illusionsconfidence}, confidence and uncertainty calibration
\cite{wang2026faithfulconfidence,xie2026knowwhenwrong,muller2026benchmarkinguq,
liu2026thinkuncertainty}, abstention-oriented interventions
\cite{wen2025abstentionSurvey,zong2026icalm}, and internal-state probes of
truthfulness \cite{ji2024_internal_states_hallu_risk,orgad2025knowmore,luo2026twopathways}.
These works ask whether a model is uncertain, inconsistent, or internally carrying
truthfulness cues. Our question is adjacent but different: when can a response be
locally stable and still fail to track truth? Prior work has also explored
connections between Kantian themes, cybernetics, and epistemic feedback
\cite{sachs2022_cybernetic_kant, burmeister2021_kant_cybernetics_security,
marlowe2021_cybernetics_philosophy}, and recent studies have analyzed instability
and hallucination in AI systems through related notions of internal model fragility
\cite{ji2024_internal_states_hallu_risk, impossibility2025_hallucination_control}.
This paper proposes a structural framework that links Kantian self-limitation to
closed-loop state-estimation operators and to simple LLM probes, treating epistemic
stability as a shared design problem rather than as a guarantee of truth.

\section{Theory: From Kant to Closed-Loop Stability}
\label{sec:theory}

This section gives the conceptual bridge used by the rest of the paper. Kant supplies the
commitment-gate idea: judgment is warranted only when the conditions for judgment are in
place. The linear model below supplies a minimal engineering setting in which stability can
be separated from truth-tracking.

\subsection{Kantian self-limitation as feedback}
Kant's \emph{Critique of Pure Reason} asks what makes cognition possible. In the familiar
tripartite picture, \emph{sensibility} supplies appearances, \emph{understanding} organizes
them under concepts, and \emph{reason} regulates understanding by enforcing systematic unity
and limiting overreach (A94/B126, A307/B364). Modern accounts of this architecture include
Allison and Guyer \cite{allison2004,guyer2006}.

We use this structure in one direction only: as an engineering analogy for feedback and
self-limitation. Sensibility provides input, understanding maintains an internal model, and
reason acts as a regulator that asks whether commitment is licensed. The point is not
doctrinal reconstruction; it is a compact design principle for abstention-aware critique.

\subsection{Minimal state-space abstraction}
\label{subsec:state-space-abstraction}

Around an operating point, the prediction-correction picture can be approximated by a
linear--Gaussian state-space model:
\begin{equation}
\begin{aligned}
x_{t+1} &= A x_t + w_t,\\
y_t &= H x_t + v_t,
\end{aligned}
\label{eq:lti-ss}
\end{equation}
where $x_t$ is the internal state, $y_t$ is the observation, and $w_t,v_t$ are process and
measurement noise with covariances $Q,R$. A Kalman-style correction updates the prediction by
\begin{equation}
\hat x_{t|t}
= \hat x_{t|t-1}
  + K_t\!\left(y_t-H\hat x_{t|t-1}\right),
\qquad
K_t=P_{t|t-1}H^\top(HP_{t|t-1}H^\top+R)^{-1}.
\end{equation}
The gain $K_t$ trades trust in the internal model against trust in incoming evidence. This
is the control-theoretic analogue of a commitment gate: the system does not simply preserve
its current state, but decides how strongly observations should correct it.

For the rest of the paper we use a time-invariant gain $K$ and define the closed-loop error
operator
\begin{equation}
\Phi \equiv A-KH.
\label{eq:phi-def}
\end{equation}
When $\rho(\Phi)<1$ and the pair $(A,H)$ is detectable, estimation error remains bounded.
But bounded error dynamics do not imply warranted judgment. A loop may be stable while
settling around the wrong state; this is the linear analogue of a false fixed point.

\subsection{Bridge to measurement}

The empirical question is therefore not whether a system is merely stable, but what kind of
stability it has. Non-normal or ill-conditioned closed loops can amplify perturbations even
when their eigenvalues are stable \cite{trefethen2005}. Conversely, a high-signal,
low-sensitivity system may be locally inert without tracking truth.

Section~\ref{sec:metrics} turns this distinction into measurable descriptors: spectral
margin, conditioning, integrated sensitivity, and innovation amplification. For LLMs we
cannot read off $\Phi$, so the experiments use output-level audit signals and hidden-state
perturbation sensitivity as proxies. The hypothesis tested later is simple: if
overconfident errors were just brittle computations, OCW items should be more locally
sensitive than CC items. The results do not support that simple gap.

\section{Measuring Epistemic Instability: H-Risk}
\label{sec:metrics}
\noindent
This section has two jobs. First, it makes the feedback-stability idea measurable in a
linear--Gaussian setting where the closed-loop operator is explicit. Second, it defines the
small LLM audit score used in the experiments. These are related but not identical:
$\Hrisk_{\mathrm{LTI}}$ is a structural index in a controlled model, where LTI means
linear time-invariant. By contrast,
$H_{\mathrm{proxy}}$ is a labeled audit triage score for deciding where to spend evaluation
and intervention effort. The relation is local: the LTI model explains why policy-wise
confidence variation and confident-wrong mass are natural audit signals, while
$H_{\mathrm{proxy}}$ measures those signals directly on labeled LLM outputs. It is an
operational probe, not the paper's central theoretical claim. We then
situate this view relative to existing output-centric
hallucination metrics (Sec.~\ref{subsec:related-metrics}).

\begin{table}[t]
  \centering
  \caption{\textbf{Three H-Risk objects used in the paper.} The table fixes the role and
  claim strength of each object before the formal definitions.}
  \label{tab:hrisk_objects}
  \small
  \setlength{\tabcolsep}{3pt}
  \begin{tabular}{p{0.15\linewidth}p{0.35\linewidth}p{0.12\linewidth}p{0.18\linewidth}}
    \toprule
    Name & Role & Measured? & Claim strength \\
    \midrule
    $\Hrisk_{\mathrm{LTI}}$ & LTI toy/control index for a known closed-loop operator &
    Yes & Formal illustration \\
    $\Hrisk_{\mathrm{LLM}}$ & Desired operator-level LLM index based on hidden-state
    dynamics & No & Future target \\
    $H_{\mathrm{proxy}}$ & Label-aware audit proxy for retrospective domain triage &
    Yes & Operational probe only \\
    \bottomrule
  \end{tabular}
\end{table}

\subsection{Abstract definition of H-Risk}
\label{subsec:abstract-hrisk}
We begin by abstracting away from any particular architecture and treating inference as a
discrete-time dynamical system with closed-loop operator $\Phi$, as in Eq.~\eqref{eq:phi-def}.
To each such operator we associate four nonnegative, dimensionless descriptors
$$(m_R,c_R,s_R,a_R)\in[0,\infty)^4,$$
representing respectively an instability margin, a conditioning factor, an integrated
sensitivity, and an innovation amplification term.

\begin{definition}[Abstract H-Risk]
\label{def:abstract-hrisk}
An abstract hallucination risk index is any normalized scalar $S(\Phi)$ that is
nondecreasing in each of $(m_R,c_R,s_R,a_R)$, separates their contributions up to monotone
rescaling, and equals $1$ at a fixed stable reference configuration $\Phi_0$.
\end{definition}

The definition fixes the direction of the scale without pretending that the product below
is unique. In this paper we use one canonical linear--Gaussian instantiation and a separate
output-level proxy for LLM audit sets.

\subsection{Linear--Gaussian instantiation}
\label{subsec:lti-hrisk}
We first instantiate H-Risk in the classical setting of the linear--Gaussian state-space 
model \eqref{eq:lti-ss}, with process and measurement noises $w_t \sim \mathcal{N}(0,Q)$ and 
$v_t \sim \mathcal{N}(0,R)$, and a steady-state Kalman filter with gain $K$.
The associated closed-loop operator on the state is
$\Phi = A-KH$,\footnote{Some conventions, depending on whether errors are
defined a priori or a posteriori, yield the equivalent form $(I-KH)A$; we use
the Luenberger form $A-KH$ throughout.} which governs how estimation errors
propagate over time.

Let
\begin{equation}
  \Sigma_e = Q + K R K^\top,
  \qquad
  P_\Phi = \Phi P_\Phi \Phi^\top + \Sigma_e,
  \label{eq:lti-error-cov}
\end{equation}
where $P_\Phi$ is the steady-state error covariance when $\rho(\Phi)<1$, and define the
Lyapunov resolvent
\begin{equation}
  M_\Phi = I_{n^2} - \Phi \otimes \Phi .
\end{equation}
In the LTI experiments we use the following concrete descriptors:
\begin{align}
  m_{\mathrm{LTI}}(\Phi)
  &= \frac{1}{1-\rho(\Phi)}, \qquad \rho(\Phi)<1, \\
  c_{\mathrm{LTI}}(\Phi)
  &= \kappa_2(\Phi)
   = \frac{\sigma_{\max}(\Phi)}{\sigma_{\min}(\Phi)}, \\
  s_{\mathrm{LTI}}(\Phi)
  &= \left\| M_\Phi^\dagger \right\|_2, \\
  a_{\mathrm{LTI}}(H,P_\Phi,R)
  &= \frac{\operatorname{tr}(H P_\Phi H^\top)}{\operatorname{tr}(R)} .
  \label{eq:lti-descriptors}
\end{align}

\begin{table}[t]
  \centering
  \caption{\textbf{Linear--Gaussian H-Risk descriptors.} Larger values mean weaker stability,
  poorer conditioning, stronger covariance sensitivity, or larger innovation energy relative
  to measurement noise.}
  \label{tab:hrisk_descriptors}
  \small
  \setlength{\tabcolsep}{4pt}
  \begin{tabular}{p{0.12\linewidth}p{0.31\linewidth}p{0.43\linewidth}}
    \toprule
    Term & Formula & Meaning \\
    \midrule
    $m_{\mathrm{LTI}}$ & $(1-\rho(\Phi))^{-1}$ & Distance to closed-loop instability \\
    $c_{\mathrm{LTI}}$ & $\sigma_{\max}(\Phi)/\sigma_{\min}(\Phi)$ & Euclidean conditioning of the update \\
    $s_{\mathrm{LTI}}$ & $\| (I-\Phi\otimes\Phi)^\dagger \|_2$ & Sensitivity of steady-state covariance \\
    $a_{\mathrm{LTI}}$ &
    $\operatorname{tr}(HP_\Phi H^\top)/\operatorname{tr}(R)$ &
    Innovation energy relative to noise \\
    \bottomrule
  \end{tabular}
\end{table}

For singular $\Phi$, $c_{\mathrm{LTI}}$ is treated as infinite.

From these raw descriptors we construct \emph{normalized} descriptors
\begin{equation}
  \bar m_{\mathrm{LTI}}=\frac{m_{\mathrm{LTI}}}{m_{\mathrm{LTI},0}},
  \quad
  \bar c_{\mathrm{LTI}}=\frac{c_{\mathrm{LTI}}}{c_{\mathrm{LTI},0}},
  \quad
  \bar s_{\mathrm{LTI}}=\frac{s_{\mathrm{LTI}}}{s_{\mathrm{LTI},0}},
  \quad
  \bar a_{\mathrm{LTI}}=\frac{a_{\mathrm{LTI}}}{a_{\mathrm{LTI},0}},
  \label{eq:lti-normalization}
\end{equation}
where the subscript $0$ denotes the fixed reference configuration used in the simulation
sweep. In the released code, this reference is the stable Kalman-filter configuration with
the same $A,Q,R$ as the sweep, $H_0=[1,\epsilon_0]$ with $\epsilon_0=0.3$, and $K_0$ equal to
the corresponding steady-state Kalman gain. We then define the linear--Gaussian H-Risk as
the product

\begin{equation}
  \Hrisk_{\mathrm{LTI}} \;=\; 
  \bar m_{\mathrm{LTI}} \cdot
  \bar c_{\mathrm{LTI}} \cdot
  \bar s_{\mathrm{LTI}} \cdot
  \bar a_{\mathrm{LTI}},
  \label{eq:hrisk-lti-def}
\end{equation}
which satisfies Definition~\ref{def:abstract-hrisk}.
We adopt $\Hrisk_{\mathrm{LTI}}$ in \eqref{eq:hrisk-lti-def} as the canonical 
linear--Gaussian instantiation of abstract H-Risk.

\paragraph{Basis and metric.}
The spectral radius term is invariant under similarity transforms, but the Euclidean
conditioning and Lyapunov-resolvent norms above are not coordinate-free. We therefore treat
$\Hrisk_{\mathrm{LTI}}$ as a diagnostic defined only after a metric and basis have been
specified. In the experiments reported here, all
singular values and operator norms are computed in the fixed two-dimensional simulation
state basis of Eq.~\eqref{eq:lti-ss}, using the standard Euclidean metric after the
state/noise scaling encoded by $A,Q,R$; no whitening transform is applied. This is adequate
for within-sweep comparisons because every configuration is expressed in the same basis and
normalized against the same reference point, but the resulting value should not be read as a
coordinate-invariant scalar. A cross-basis comparison would require a predeclared whitening
map, for example replacing $(\Phi,H,\Sigma_e)$ by $(W\Phi W^{-1},HW^{-1},W\Sigma_e W^\top)$
before computing the norm-based terms.

\subsection{LLM instantiation of H-Risk}
To apply H-Risk to large language models, we conceptually view generation as a recurrent 
update on a high-dimensional hidden state $h_t$, with a token sequence $y_{1:T}$ produced 
from the evolving state. Let $J_t = \partial h_t / \partial h_{t-1}$ denote the local 
Jacobian of the hidden representation with respect to its previous context at generation 
step $t$. We then define dimensionless components
\begin{align}
  m_{\mathrm{LLM}} &= f_m(\{J_t\}_t), \\
  c_{\mathrm{LLM}} &= f_c(\{J_t\}_t), \\
  s_{\mathrm{LLM}} &= f_s(\{J_t\}_t), \\
  a_{\mathrm{LLM}} &= f_a(\{p_t,\tilde p_t\}_t),
\end{align}
where $f_m,f_c,f_s$ summarise the local Jacobians (e.g., stability margins, non-normal 
conditioning measures, and temporal sensitivity norms), and $f_a$ compares token-level 
innovation statistics to calibrated uncertainty estimates derived from token probabilities 
$p_t$ and auxiliary critic distributions $\tilde p_t$.

After choosing application-specific normalizations, we construct normalized descriptors 
$\bar m_{\mathrm{LLM}}, \bar c_{\mathrm{LLM}}, \bar s_{\mathrm{LLM}}, \bar a_{\mathrm{LLM}}$ 
such that each equals $1$ for a reference, well-calibrated model and prompt regime. 
Formally, this suggests a notional operator-level index

\begin{equation}
  \Hrisk_{\mathrm{LLM}} \;=\; 
  \bar m_{\mathrm{LLM}} \cdot
  \bar c_{\mathrm{LLM}} \cdot
  \bar s_{\mathrm{LLM}} \cdot
  \bar a_{\mathrm{LLM}},
\end{equation}
which would be an LLM-specific instantiation of abstract H-Risk in the sense of
Definition~\ref{def:abstract-hrisk}. In practice, the exact choices of $f_m,f_c,f_s,f_a$
depend on the available access to model internals and calibration signals, and typical
API-based settings do not expose the Jacobians $\{J_t\}_t$. The experiments therefore use
the audit proxy below, while direct estimation of $\Hrisk_{\mathrm{LLM}}$ remains an
operator-level target for future work.

\subsection{Output-level audit proxy for LLM audit sets}
\label{subsec:hproxy-def}

The LLM experiment uses an output-level audit proxy. Operationally,
$H_{\mathrm{proxy}}$ asks a practical question: given a labeled audit set,
which domains should receive abstention-aware evaluation, baseline comparison,
or intervention design? For each item $i$ in domain $d$ and condition
$c\in\{\mathrm{C0},\mathrm{C1},\mathrm{C2}\}$, let $p_{ic}$ denote the
policy's self-reported confidence that its current action is correct. In the recollected
C0/C1/C2 audit used here, answered items use the reported
$P(\mathrm{correct})\in[0,1]$; abstentions are assigned the neutral value $p_{ic}=0.5$.
We use this verbal confidence as an audit signal produced by the policy run.

Let $\mathcal{C}_i$ be the observed subset of $\{\mathrm{C0},\mathrm{C1},\mathrm{C2}\}$ for
item $i$, and keep items with $|\mathcal{C}_i|\ge 2$. The output-level policy-variation term
is the population standard deviation
\begin{equation}
  V_i
  =
  \operatorname{Std}_{c\in\mathcal{C}_i}
  \left[p_{ic}\right],
  \qquad \text{with abstentions coded as } p_{ic}=0.5 .
  \label{eq:hproxy-variation}
\end{equation}
The label-aware overconfident-wrong term is computed only from the forced-answer baseline:
\begin{equation}
  O_i
  =
  \mathbb{I}\{\mathrm{answered}_{i,\mathrm{C0}}\}
  \mathbb{I}\{p_{i,\mathrm{C0}}\ge p_{\mathrm{hi}}\}
  \mathbb{I}\{y_{i,\mathrm{C0}}=0\},
  \qquad p_{\mathrm{hi}}=0.8 ,
  \label{eq:hproxy-ocw}
\end{equation}
where $y=1$ denotes a correct answer and $y=0$ an incorrect answer. The per-item audit score
and domain aggregation are
\begin{align}
  h_i &= V_i + \lambda O_i, \qquad \lambda=1, \\
  H_{\mathrm{proxy}}(d)
  &= \frac{1}{|\mathcal{I}_d|}
     \sum_{i\in\mathcal{I}_d} h_i, \\
  H_{\mathrm{proxy}}^{\mathrm{norm}}(d)
  &= \frac{H_{\mathrm{proxy}}(d)}
          {\max_{d'} H_{\mathrm{proxy}}(d')}.
  \label{eq:hproxy-domain}
\end{align}
The inputs are therefore a labeled audit table with domain, item identifier, condition,
answer/refusal status, policy confidence, and observed correctness. The normalization in
\eqref{eq:hproxy-domain} is used only for the domain-ranking plot; the unnormalized mean and
within-domain standard deviation are retained as descriptive summaries.

$H_{\mathrm{proxy}}$ is label-aware because the $O_i$ term requires knowing whether C0 is
wrong. It is therefore a retrospective audit score for labeled evaluation sets, not a
deployment-time uncertainty estimator for unlabeled inputs. Nor is it meant to beat direct
label-aware failure summaries such as C0 error rate or C0 Brier risk as a pure ranker. Its
role is interpretive: it decomposes audited domains into policy-level confidence movement
and overconfident-wrong mass, so that an audit can ask whether a domain is merely hard or
specifically exposed to abstention-aware policy changes.

\paragraph{Relation to the LTI index.}
The LTI construction supplies an explanatory bridge; it does not derive
$H_{\mathrm{proxy}}$ as a unique score. In the LTI setting, the structural terms in
$\Hrisk_{\mathrm{LTI}}$ describe how a closed loop reacts to small changes in its correction
rule and observations. If we add a local family of nearby stabilising gains and a binary
readout, these same factors have observable consequences: confidence can vary across nearby
policies, and a reference policy can be confident but wrong. The LLM proxy keeps only these
observable consequences. Its variation term is the finite-policy analogue of confidence
spread under nearby corrections; its OCW term is the labeled analogue of a high-confidence
wrong reference decision. Thus $H_{\mathrm{proxy}}$ translates the LTI lesson into a
domain-level audit question for settings where $\Phi$ and Jacobians are unavailable. The
bridge is local and low-order, and its role is to motivate observable audit signals rather
than to provide a mechanistic reduction of LLM inference.

\subsection{Related hallucination metrics and their limitations}
\label{subsec:related-metrics}
Recent work proposes a spectrum of output-level hallucination and consistency metrics for
large language models. Surveys and taxonomies now provide overviews of hallucination types,
causes, detection, and mitigation \cite{ji2023survey,alansari2025hallu_survey}. On the
detection side, proposed metrics range from self-consistency based disagreement and
factuality scores, as in early black-box detectors such as SelfCheckGPT
\cite{manakul2023selfcheckgpt}, to semantic uncertainty methods such as semantic entropy
and semantic energy \cite{farquhar2024semanticentropy,ma2025semanticenergy}, domain-specific
benchmarks such as Molecular Mirage for scientific hallucinations
\cite{li2025molecular_mirage}, and black-box measures based on consistency under uncertain
expressions \cite{joo2025consistency}. Newer work also stresses robustness beyond pointwise
confidence: neighborhood consistency probes whether beliefs persist under contextual
interference \cite{xu2026illusionsconfidence}, certainty-robustness benchmarks test whether
answers are stable under conversational challenge \cite{saadat2026certaintyrobustness}, and
long-form QA benchmarks show that verbal or token-level uncertainty can be unreliable in
reasoning-heavy settings \cite{muller2026benchmarkinguq}. Re-evaluation work has also
questioned how robust headline gains in hallucination detection really are, showing that
apparent progress can depend sensitively on the choice of metric and benchmark
\cite{janiak2025illusion}. These contributions are valuable, but most remain
\emph{output-facing}: they evaluate whether the final text, sampled meanings, or neighboring
answers align with external sources or majority judgments. From our perspective, they
motivate rather than replace the central separation: local robustness, surface consistency,
and truth-tracking are related but non-identical properties.

Prompt-based critique-and-revision (CnR) methods such as Self-Refine, Reflexion, CRITIC, and
related approaches (e.g., \cite{madaan2023selfrefine,shinn2023reflexion,gou2023critic,
chan2023chateval}) introduce an explicit feedback loop: an initial answer is critiqued and
then revised, often with an auxiliary LLM-as-judge module or multi-agent debate. A parallel
line studies abstention and selective answering, including surveys of abstention behavior,
risk-sensitive confidence policies, and prompt-only abstention frontiers
\cite{wen2025abstentionSurvey,wang2026faithfulconfidence,zong2026icalm,
liu2026thinkuncertainty}. Our framing places these methods within the same broad design
space but shifts the emphasis from improving an answer to checking whether commitment is
warranted. In this sense, $\Hrisk$ and $H_{\mathrm{proxy}}$ complement output-level metrics and
CnR-style procedures rather than competing with them, supplying a structural notion of
epistemic stability that can be monitored alongside task-level performance.

\section{Experimental Setup and Internal Probes}
\label{sec:method}

\paragraph{Data, domains, and pairing.}
We study short, single-turn binary factual items grouped into $11$ topical domains. In the
recollected proxy study, all three policies (C0/C1/C2) are available on the same
$N=532$ items, so paired $\Delta SE_{\mathrm{policy}}$ comparisons use the full shared item
set. Domain sizes range from $35$ to $64$ items. The item file was frozen before policy
evaluation. Binary gold labels were assigned independently of model outputs, and items with
missing or ambiguous written truth conditions were excluded from the frozen audit set. Domain
labels are used only for aggregation and are not shown to the model.

\paragraph{Policies.}
We compare three policies in the frozen main audit. C0 is a forced-answer baseline that
must output \texttt{Yes.} or \texttt{No.}; it is forced judgment. C1 is cautious
judgment: it allows abstention only when the question is too under-specified to answer
responsibly. C2 is self-critical judgment: a single-shot, risk-aware policy that explicitly
checks for uncertainty and may abstain. C2 is therefore a lightweight implementation of the
commitment-gate idea, not a full multi-step feedback loop.

\paragraph{Operational feedback protocol.}
To make the feedback concept executable beyond the frozen C0--C2 run, the released
evaluation script also specifies C3, an explicit loop with four internal steps: initial
answer, warrant critique, revision or abstention gate, and final confidence report. This
is feedback-gated judgment in the operational sense: the model first proposes an answer,
then checks whether the warrant for commitment is sufficient.
Because these logs were not part of the frozen main C0--C2 audit, we report C3 only as a
post-hoc $N=100$ pilot on paired frozen-audit items in Section~\ref{subsec:llm-results}.

We also run a stricter C3-R follow-up on all $532$ frozen audit items. C3-R is designed to
make the commitment gate less discretionary: the model must report predeclared hard
blockers for vague/non-resolvable comparisons, unavailable current or private data, concrete
defeaters, answer instability after critique, and failure to state what evidence would make
the answer true or false. Offline, the policy commits only when no blocker fires and the
reported post-critique confidence exceeds a threshold. The threshold is selected on a
deterministic dev split, then applied once to the held-out split; the test set is not used
to choose the threshold. We report both a utility-selected threshold and a coverage-matched
threshold, so the comparison cannot silently buy improvement by tuning coverage on the test
items.

\paragraph{Output-level policy model.}
All C0--C2 policy logs were collected with \texttt{gpt-4.1-mini}. Each policy returns both a
decision and a verbal $P(\mathrm{correct})$ score; the gold label is never included in the
prompt. We use this verbal confidence as the policy's reported confidence signal.
Exact replay may depend on the hosted model snapshot, so the analysis uses the
frozen policy-output CSV and derived aggregates.

\paragraph{Empirical instantiation of $H_{\mathrm{proxy}}(d)$.}
Section~\ref{subsec:hproxy-def} defines the audit proxy. The empirical implementation uses
the frozen C0/C1/C2 output table, codes abstentions as confidence $0.5$, marks C0
high-confidence wrong answers at threshold $0.8$, and averages the resulting item scores
within each domain. Algorithm~\ref{alg:hproxy} records the computation once; later sections
refer back to this definition.

\begin{algorithm}[H]
\caption{Computing $H_{\mathrm{proxy}}(d)$ from policy outputs}
\label{alg:hproxy}
\begin{algorithmic}[1]
\Require Items in domain $d$; policies C0, C1, C2; one confidence value per item and policy;
high-confidence threshold $p_{\mathrm{hi}}$; weight $\lambda$.
\State If a policy abstains on an item, replace its confidence by $0.5$; otherwise use the
reported $P(\mathrm{correct})$.
\For{each item $i$ in domain $d$}
\State Compute the confidence-variation term as the standard deviation across the three
policy confidences.
\State Set the high-confidence-wrong flag to $1$ if C0 answers, its confidence is at least
$p_{\mathrm{hi}}$, and C0 is wrong; otherwise set it to $0$.
\EndFor
\State \textbf{Output:} average over domain items of
``confidence variation $+$ $\lambda\times$ high-confidence-wrong flag''.
\end{algorithmic}
\end{algorithm}

\paragraph{Policy-aware loss.}
For the abstention-aware evaluation, let $y_i\in\{0,1\}$ denote the binary gold label for
the positive \texttt{Yes.} answer. We convert each policy action into the probability assigned
to \texttt{Yes.}:
\[
\tilde p_{i,c} =
\begin{cases}
P(\mathrm{correct}), & \text{if the policy answers \texttt{Yes.}},\\
1-P(\mathrm{correct}), & \text{if the policy answers \texttt{No.}},\\
0.5, & \text{if the policy abstains.}
\end{cases}
\]
The policy-aware squared loss is
\begin{equation}
  SE_{\mathrm{policy}}(i,c) = (\tilde p_{i,c}-y_i)^2 .
  \label{eq:se-policy}
\end{equation}
In the default analysis an abstention receives the neutral loss $(0.5-y_i)^2=0.25$. Because
this is a utility choice rather than a fact about correctness, we also sweep the abstention
loss from $0.05$ to $0.50$ in Section~\ref{subsec:llm-results}. We compare C1 and C2 to the
forced-answer baseline item by item using
\[
  \Delta SE_{\mathrm{policy}}(i,c)
  =
  SE_{\mathrm{policy}}(i,c)-SE_{\mathrm{policy}}(i,\mathrm{C0}),
\]
where negative values indicate reduced loss.

\paragraph{Internal sensitivity probe.}
To complement the output-level proxy, we measure how much hidden states move when we make a
small change to the input embeddings. Let $e\in\mathbb{R}^{T\times D}$ be the input
embedding sequence for one item, and draw a Gaussian perturbation
$\epsilon\sim\mathcal{N}(0,\sigma^2 I)$ of the same shape. For a chosen layer $\ell$ and,
by default, the final token, let $h_{\ell}^{\mathrm{clean}}$ be the hidden state from the
clean input $e$, and let $h_{\ell}^{\mathrm{noisy}}$ be the hidden state from the perturbed
input $e+\epsilon$. We define the per-trial local sensitivity as
\begin{equation}
  S_\ell(\epsilon)
  =
  \frac{\|h_\ell^{\mathrm{noisy}}-h_\ell^{\mathrm{clean}}\|_2}
       {\|e^{\mathrm{noisy}}-e^{\mathrm{clean}}\|_2}.
  \label{eq:sl-def}
\end{equation}
We estimate $S_\ell$ by averaging over $n_{\mathrm{trials}}=40$ perturbations per item at
noise scale $\sigma=0.01$, then report the mean and standard error over items. The internal
probe uses greedy decoding with sampling disabled and compares standard versus self-critical
prompt variants on the same items. We probe four representative depths per model.

\paragraph{Jacobian proxy for $\Phi_{\mathrm{LLM}}$ (conceptual).}
In the LTI toy model, local sensitivity is controlled by a linear operator. In an LLM, the
closest local analogue would be a Jacobian of the hidden state with respect to its context.
Explicit Jacobian estimation at modern model widths and sequence lengths is expensive, so
the hidden-state sensitivity above serves as the practical local approximation used in this
paper.

\FloatBarrier

\section{Results}
\label{sec:results}

We summarise the LLM probes in three steps.
First, we test the core false-fixed-point question directly: whether overconfidently wrong
items are more locally fragile than confidently correct ones.
Second, we examine how abstention-aware commitment policies trade coverage for fewer
overconfident false factual judgments, and how a small retrospective audit proxy ranks
domains where that tradeoff is most relevant.
Third, we report a C3-R follow-up that turns the commitment gate into an explicit
post-hoc safety policy.

\subsection{Stabilization of internal representations}
\label{subsec:internal-stability}

We begin with the central empirical question: are high-confidence false factual judgments
locally more fragile than confidently correct ones?
Using the sensitivity probe defined in Section~\ref{sec:method}, we compare standard and
self-critical instruction variants on the same items for Llama-3.1-8B-Instruct,
DeepSeek-R1-Distill-Llama-8B, and Qwen2.5-7B-Instruct. Lower $S_\ell$ means less
hidden-state movement under the fixed $\sigma=0.01$ embedding perturbation.

Across all three models, the self-critical prompt lowers mean local sensitivity across depth.
To test whether overconfident errors coincide with an internal instability gap, we split C0
items into confidently correct (CC) and overconfidently wrong (OCW) groups and compare their
final-layer sensitivities under the standard prompt. Table~\ref{tab:sl_gap_cc_ocw} shows that
the relative OCW--CC differences are small: $+2.2\%$ for Llama-3.1, $+6.1\%$ for DeepSeek-R1,
and $+0.4\%$ for Qwen2.5. With a practical equivalence margin of $\pm10\%$, the bootstrap
intervals lie within the margin for Llama-3.1 and Qwen2.5; DeepSeek-R1 remains
inconclusive because its upper interval exceeds the margin. Under this local Gaussian probe,
we therefore find no systematic OCW-specific fragility gap.
This is the main empirical reason to take false fixed points seriously: the wrong answers
are not simply the answers that move most easily.

\begin{figure}[htbp]
  \centering
  \includegraphics[width=\textwidth]{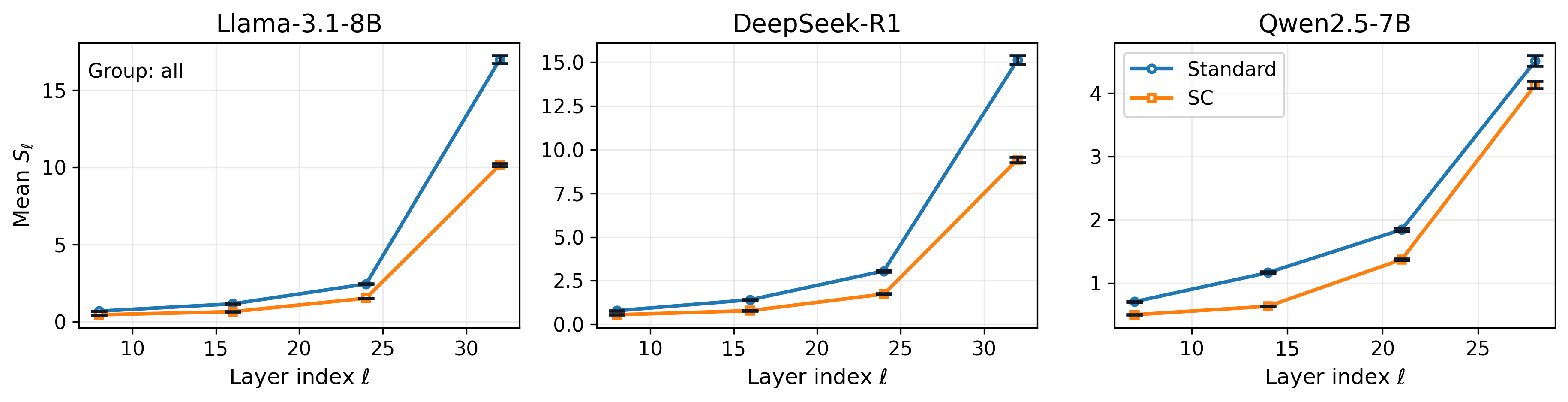}
  \caption{Layer-wise internal sensitivity $S_\ell$ under standard vs self-critical prompts
  for three open-weight instruction-tuned models. Self-critique reduces mean local
  sensitivity across depth. The CC--OCW sensitivity gap under the standard prompt is
  small relative to the absolute sensitivity scale, with equivalence outcomes reported in
  Table~\ref{tab:sl_gap_cc_ocw}.
  Error bars show standard errors over items.}
  \label{fig:internal_sensitivity_1x3}
\end{figure}

\begin{table}[H]
  \centering
  \caption{Final-layer local sensitivity $S_{\ell}$ for confidently correct (CC) versus
  overconfidently wrong (OCW) groups under the standard prompt (C0; $\sigma=0.01$).
  Relative differences are small. Equivalence is assessed using a practical $\pm10\%$
  margin and bootstrap confidence intervals over items.}
  \label{tab:sl_gap_cc_ocw}
  \begin{adjustbox}{max width=\linewidth}
    \begin{tabular}{lrrrr}
\toprule
Model & CC & OCW & Rel. diff 95\% CI & Within $\pm10\%$? \\
\midrule
Llama-3.1-8B & 16.48 & 16.84 & +2.2\% [-2.8, +7.5] & Yes \\
DeepSeek-R1 & 14.62 & 15.50 & +6.1\% [-1.0, +13.2] & Inconclusive \\
Qwen2.5-7B & 4.53 & 4.55 & +0.4\% [-5.9, +6.9] & Yes \\
\bottomrule
\end{tabular}

  \end{adjustbox}
\end{table}

\FloatBarrier

\subsection{Operational audit probe with abstention-aware evaluation}
\label{subsec:llm-results}

\begin{figure}[t]
  \centering
  \includegraphics[width=\textwidth]{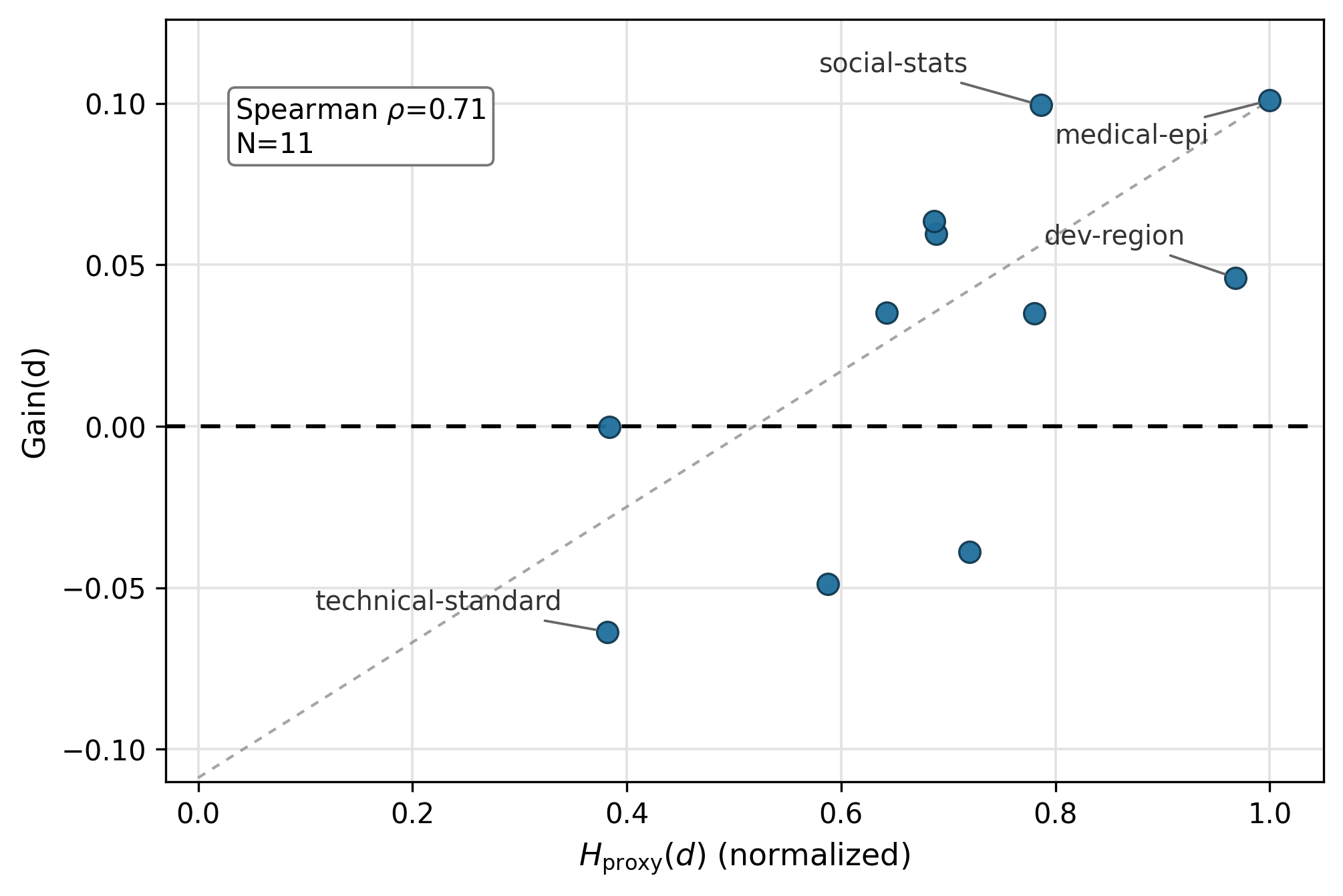}
  \caption[Proxy correlates with domain-wise gains]{\textbf{Proxy correlates with
  domain-wise policy gains.} Normalised $H_{\mathrm{proxy}}(d)$ versus C2 gain over C0,
  where $\mathrm{Gain}(d)=SE_{\mathrm{policy}}(d,\mathrm{C0})-
  SE_{\mathrm{policy}}(d,\mathrm{C2})$ and higher is better. The dashed line is a Theil--Sen
  fit for visualization only; the ranking summary is Spearman $\rho=0.71$ over $11$
  domains, with 95\% bootstrap CI $[0.14,0.97]$.}
  \label{fig:proxy_predicts_gain}
\end{figure}

The practical question is where extra caution is worth paying for.
Here the score has a narrow role: it connects the commitment-gate idea to audited policy
outputs.
In the recollected $N=532$ audit set, $H_{\mathrm{proxy}}(d)$ decomposes domains where
policy-level confidence variation and overconfident-wrong mass coincide.
Figure~\ref{fig:proxy_predicts_gain} plots
normalized $H_{\mathrm{proxy}}(d)$ against the C2 gain over C0,
\[
  \mathrm{Gain}(d)
  =
  SE_{\mathrm{policy}}(d,\mathrm{C0})
  -
  SE_{\mathrm{policy}}(d,\mathrm{C2}),
\]
so larger values are better. Across the 11 domains, the rank association is positive
(Spearman $\rho=0.71$). Medical epidemiology and social statistics show the strongest C2
reductions, while lower-gain domains such as technical standard and literature/media do not
benefit from C2 in this audit. This is the retrospective audit use specified in
Section~\ref{subsec:hproxy-def}: it identifies where extra evaluation or intervention
design is most likely to matter.

\begin{table}[t]
  \centering
  \caption{\textbf{Domain-ranking baselines.} Spearman correlations compare each domain
  score with C2 gain over C0 across 11 domains. Predictive entropy is computed from C0's
  reported confidence. Label-aware rows use observed correctness and belong to the same
  retrospective audit setting as $H_{\mathrm{proxy}}$.}
  \label{tab:uncertainty_baselines}
  \begin{adjustbox}{max width=\linewidth}
    \begin{tabular}{lcc}
\toprule
Domain score & Uses labels? & Spearman $\rho$ \\
\midrule
Predictive entropy (C0) & No & $-0.07$ $[-0.72, 0.73]$ \\
Confidence variation only & No & $-0.04$ $[-0.65, 0.64]$ \\
OCW rate only & Yes & $0.71$ $[0.18, 0.97]$ \\
$H_{\mathrm{proxy}}$ & Yes & $0.71$ $[0.12, 0.97]$ \\
C0 error rate & Yes & $0.85$ $[0.52, 1.00]$ \\
C0 Brier risk & Yes & $0.88$ $[0.55, 1.00]$ \\
\bottomrule
\end{tabular}

  \end{adjustbox}
\end{table}

Table~\ref{tab:uncertainty_baselines} makes the baseline comparison explicit. Predictive
entropy and confidence variation alone do not rank the domains where C2 helps. Label-aware
C0 error and Brier risk rank them strongly, as expected, because they measure observed
failures directly; in this small domain-level comparison they are stronger pure rankers
than $H_{\mathrm{proxy}}$. We therefore do not claim that $H_{\mathrm{proxy}}$ is the best
predictor of C2 gain. Its value is interpretive: it reports whether high-confidence
baseline errors coincide with policy movement, instead of collapsing the audit to one
labeled outcome score.

\begin{table}[t]
  \centering
  \caption{\textbf{$H_{\mathrm{proxy}}$ weight sensitivity.} The proxy is
  $V_i+\lambda O_i$, where $V_i$ is policy-level confidence variation and $O_i$ is the
  label-aware overconfident-wrong flag. The ranking is weak for variation alone
  ($\lambda=0$) and strongest near the default $\lambda=1$.}
  \label{tab:hproxy_lambda_sweep}
  \begin{adjustbox}{max width=0.62\linewidth}
    \begin{tabular}{rr}
\toprule
$\lambda$ in $V_i+\lambda O_i$ & Spearman $\rho$ with C2 gain \\
\midrule
0.00 & $-0.04$ $[-0.65, 0.64]$ \\
0.25 & $0.58$ $[-0.05, 0.94]$ \\
0.50 & $0.69$ $[0.13, 0.96]$ \\
1.00 & $0.71$ $[0.13, 0.97]$ \\
2.00 & $0.67$ $[0.07, 0.97]$ \\
\bottomrule
\end{tabular}

  \end{adjustbox}
\end{table}

As a small leakage check on this domain ranking, Table~\ref{tab:hproxy_heldout_split}
computes $H_{\mathrm{proxy}}$ on a random half of each domain and compares it with the C2
gain on the held-out half, repeating the split $2000$ times. The check is intentionally
modest: with only 11 domains it cannot establish a general predictor. The mean and median
split correlations remain positive, while the interval is wide and includes near-zero values.
We therefore treat it as a stability check for the retrospective ranking, not as a held-out
deployment validation.

\begin{table}[t]
  \centering
  \caption{\textbf{Held-out split check for $H_{\mathrm{proxy}}$.} Within each domain,
  $H_{\mathrm{proxy}}$ is computed on one random half and C2 gain is measured on the other
  half; Spearman correlations are then computed across domains. This is a stability check
  for the retrospective audit ranking, not a deployment-time validation.}
  \label{tab:hproxy_heldout_split}
  \begin{adjustbox}{max width=0.76\linewidth}
    \begin{tabular}{lrrr}
\toprule
Split check & Mean $\rho$ & Median $\rho$ [95\% interval] & Splits \\
\midrule
within-domain half split & 0.40 & 0.41 [-0.05, 0.78] & 2000 \\
\bottomrule
\end{tabular}

  \end{adjustbox}
\end{table}

\begin{figure}[t]
  \centering
  \includegraphics[width=\textwidth]{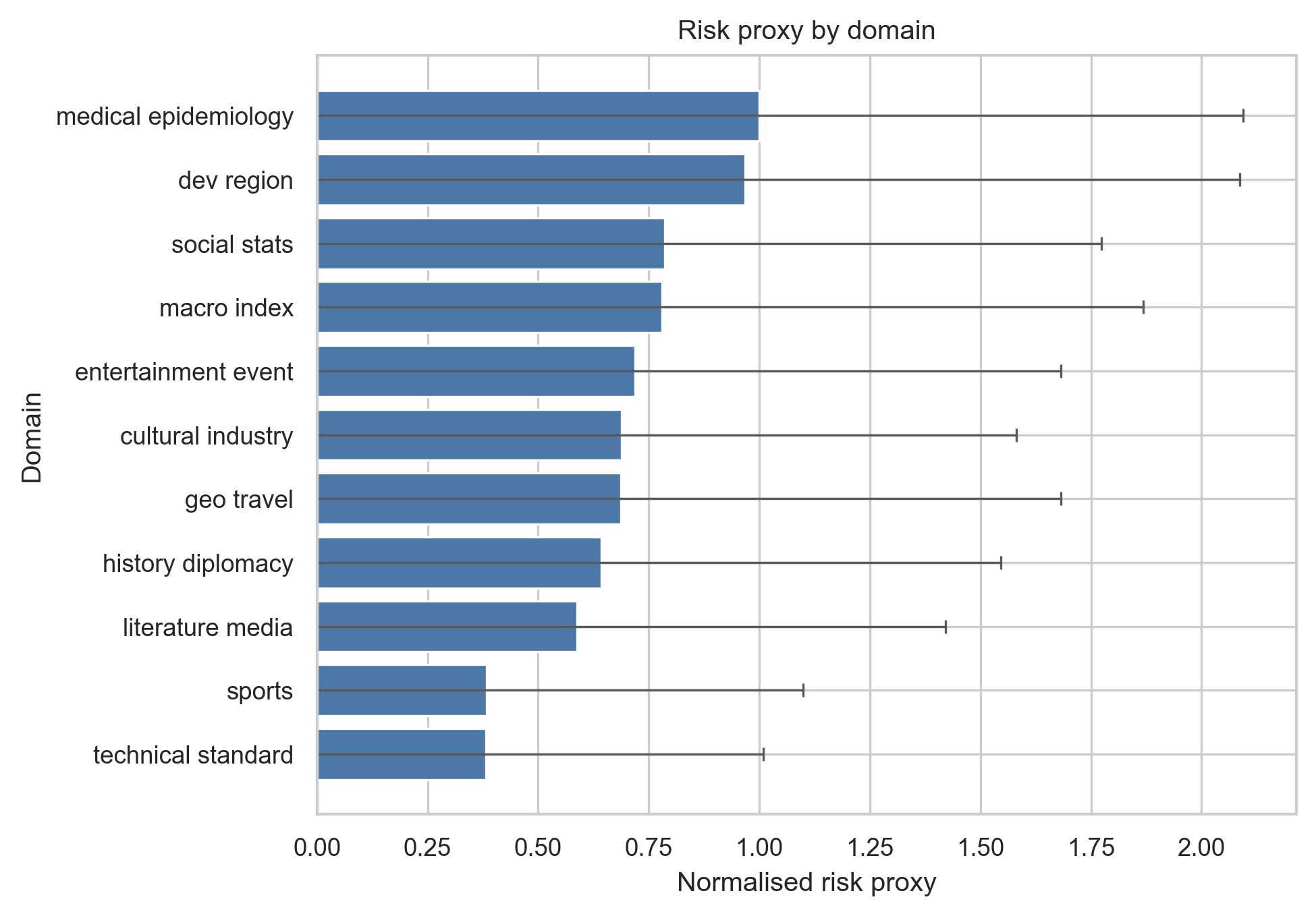}
  \caption[Domain-wise audit proxy]{\textbf{Domain-wise labeled audit proxy.} Mean per-item
  proxy $H_{\mathrm{proxy}}(d)$ by domain, normalized by the maximum across domains. Error
  bars show within-domain standard deviation and are descriptive only. Higher values indicate
  audited domains where policy-wise confidence variation and observed overconfident mistakes
  coincide.}
  \label{fig:hrisk_by_domain}
\end{figure}

\begin{figure}[t]
  \centering
  \includegraphics[width=\textwidth]{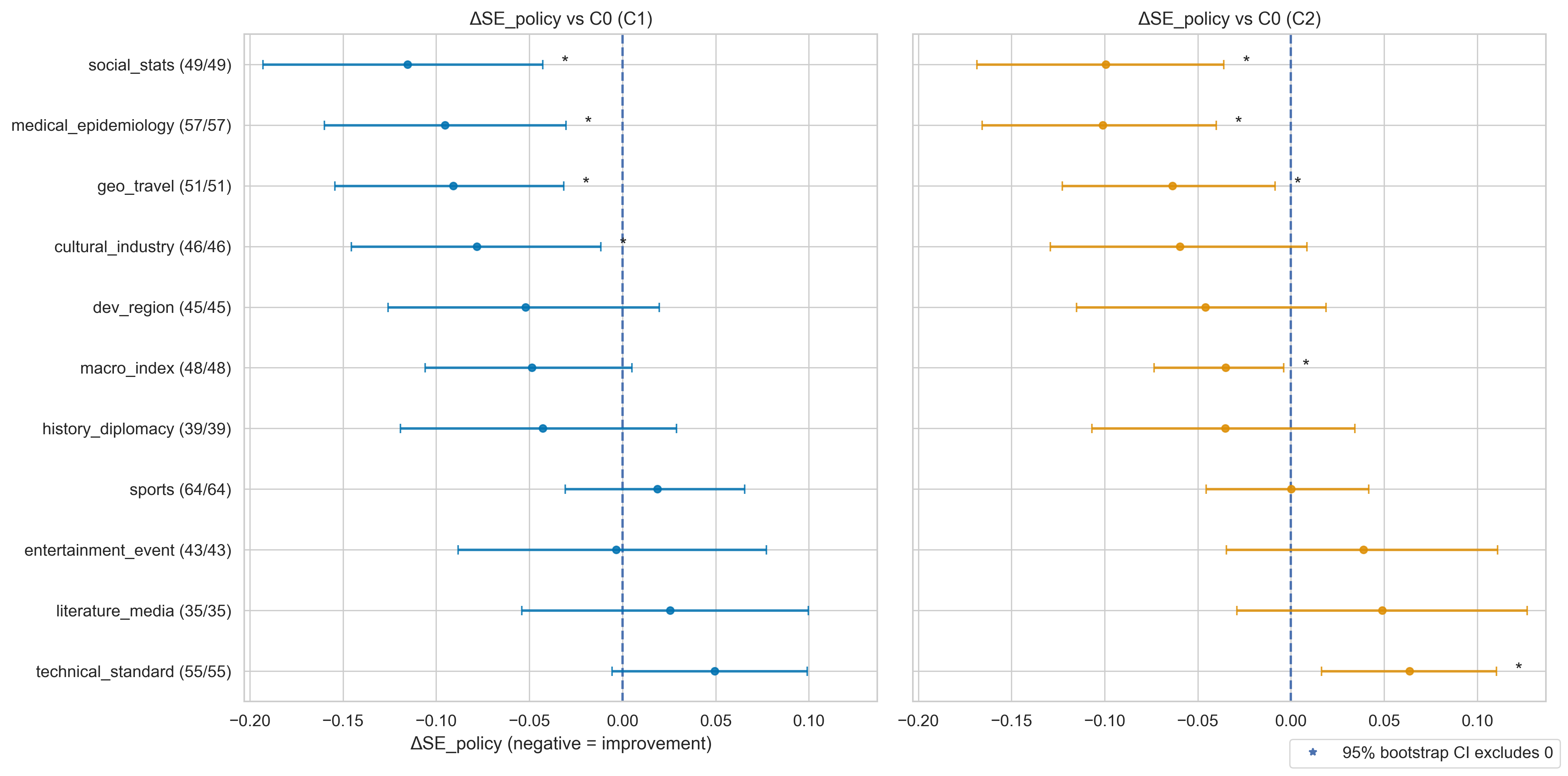}
  \caption[Policy-aware loss change by domain]{\textbf{Policy-aware loss change vs C0 by
  domain.} Mean paired per-item $\Delta SE_{\mathrm{policy}}$ (condition $-$ C0) within each
  domain; bars show 95\% paired bootstrap confidence intervals over items. Negative values
  indicate reduced policy-aware loss.}
  \label{fig:delta_brier_domain}
\end{figure}

Figure~\ref{fig:hrisk_by_domain} shows the same proxy by domain, and
Figure~\ref{fig:delta_brier_domain} shows the paired change in policy-aware squared loss for
C1 and C2 relative to the forced-answer baseline. The benefit is domain-dependent. C2 helps
most in medical epidemiology ($-0.101$), social statistics ($-0.099$), and geo travel
($-0.064$), but it is worse than C0 in entertainment event ($+0.039$), literature/media
($+0.049$), and technical standard ($+0.064$).

To make abstentions explicit, we report selective metrics in
Table~\ref{tab:selective_by_condition}. Coverage is the answer rate; selective accuracy is
accuracy conditional on answering; selective risk is one minus selective accuracy; answer
yield is overall accuracy with abstentions counted as incorrect; and OC-Wrong is the overall
rate of high-confidence mistakes. Relative to C0, both abstention-enabled policies sharply
reduce overconfident-wrong answers (from $0.211$ to $0.028$ for C1 and $0.064$ for C2), but
they do so by answering less often. C2 recovers more coverage and answer yield than C1,
while C1 attains the best conditional accuracy among answered items.

\begin{table}[t]
  \centering
  \caption{Selective metrics by condition. Coverage captures abstention; selective accuracy
  is accuracy conditional on answering; answer yield counts abstentions as incorrect; and
  OC-Wrong is the overall rate of high-confidence mistakes.}
  \label{tab:selective_by_condition}
  \begin{adjustbox}{max width=\linewidth}
    \begin{tabular}{lrrrrr}
\toprule
Condition & Coverage & Sel. Acc. & Sel. Risk & Answer Yield & OC-Wrong (overall) \\
\midrule
C0 & 1.000 & 0.639 & 0.361 & 0.639 & 0.211 \\
C1 & 0.427 & 0.744 & 0.256 & 0.318 & 0.028 \\
C2 & 0.571 & 0.674 & 0.326 & 0.385 & 0.064 \\
\bottomrule
\end{tabular}
  \end{adjustbox}
\end{table}

\begin{figure}[t]
  \centering
  \includegraphics[width=0.62\linewidth]{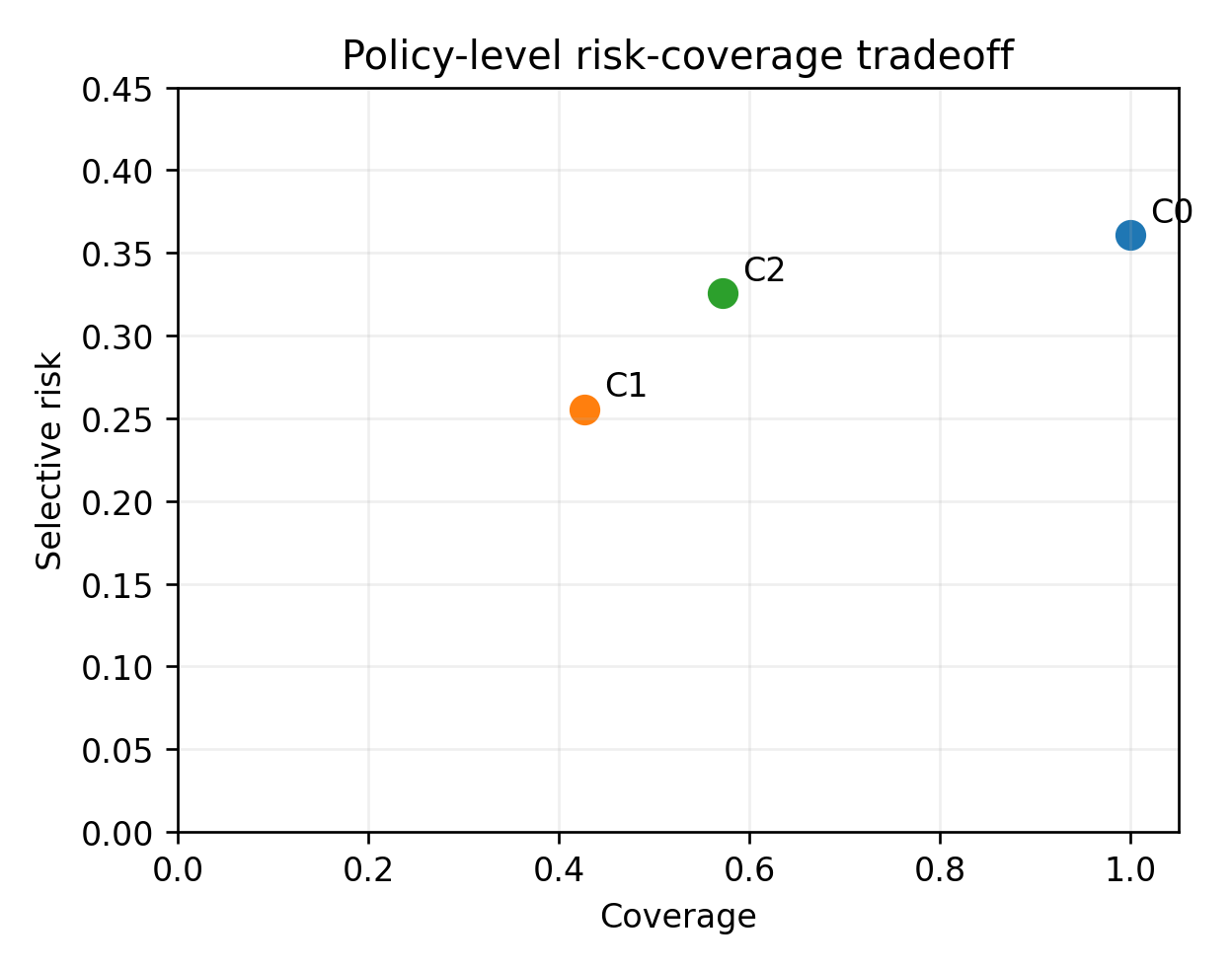}
  \caption{\textbf{Policy-level risk--coverage tradeoff.} C1 and C2 lower selective risk
  among answered items only by moving leftward in coverage. This is a policy frontier over
  three fixed prompts, not a threshold-swept risk--coverage estimate.}
  \label{fig:risk_coverage_points}
\end{figure}

\begin{figure}[t]
  \centering
  \includegraphics[width=0.82\linewidth]{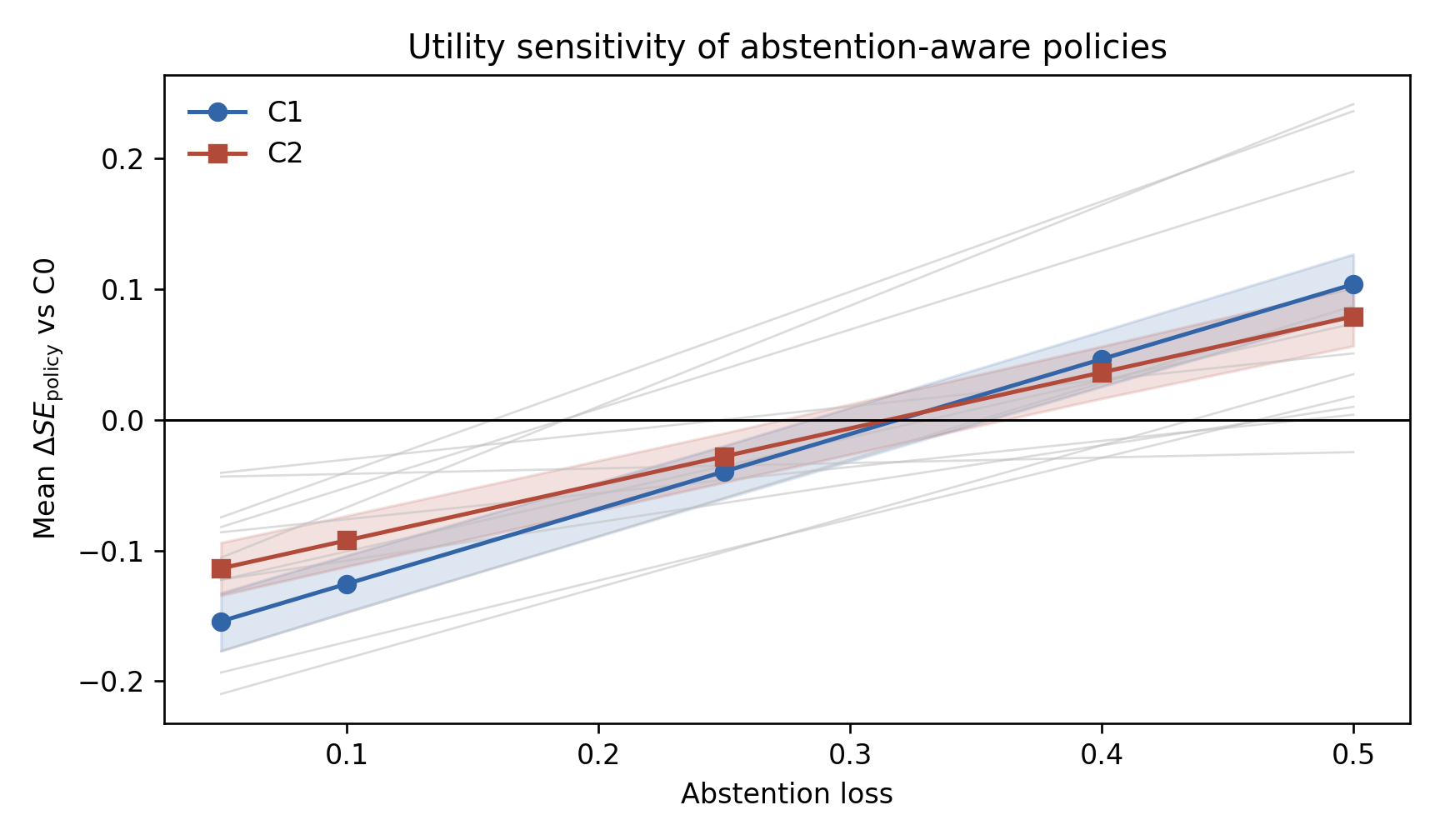}
  \caption{\textbf{Utility sensitivity to abstention loss.} Lines show paired mean
  $\Delta SE_{\mathrm{policy}}$ relative to C0 as the abstention loss varies. Thin grey
  lines show domain-level C2 deltas. The C1/C2 advantage holds when abstention is cheap, is
  still present at the default neutral loss $0.25$, and reverses when abstention is penalized
  heavily.}
  \label{fig:abstention_penalty_sweep}
\end{figure}

\begin{table}[t]
  \centering
  \caption{\textbf{Abstention-utility sweep.} Negative values mean lower policy-aware
  squared loss than C0. The default analysis uses abstention loss $0.25$.}
  \label{tab:abstention_penalty_sweep}
  \begin{adjustbox}{max width=0.82\linewidth}
    \begin{tabular}{lrrrr}
\toprule
Abstention loss & Policy & Mean $\Delta SE$ & 95\% CI & $n$ \\
\midrule
0.05 & C1 & -0.154 & [-0.177, -0.132] & 532 \\
0.05 & C2 & -0.114 & [-0.134, -0.094] & 532 \\
0.10 & C1 & -0.125 & [-0.147, -0.104] & 532 \\
0.10 & C2 & -0.092 & [-0.112, -0.073] & 532 \\
0.25 & C1 & -0.039 & [-0.060, -0.020] & 532 \\
0.25 & C2 & -0.028 & [-0.048, -0.010] & 532 \\
0.40 & C1 & 0.047 & [0.026, 0.068] & 532 \\
0.40 & C2 & 0.036 & [0.016, 0.056] & 532 \\
0.50 & C1 & 0.104 & [0.081, 0.127] & 532 \\
0.50 & C2 & 0.079 & [0.057, 0.101] & 532 \\
\bottomrule
\end{tabular}

  \end{adjustbox}
\end{table}

These results clarify the mechanism of the score-level improvement.
C1 and C2 are not generic correctness prompts; they change the judgment policy.
They trade coverage for reduced overconfident wrong answers, and under the chosen default
abstention loss this reduces policy-aware loss on average.

\subsection{Explicit C3-R feedback gate follow-up}
\label{subsec:c3r-followup}

To make the feedback-gate claim executable rather than merely definitional, we ran a
stricter C3-R follow-up on all $532$ frozen audit items. C3-R requires the model to report
predeclared warrant blockers and applies the final commit/abstain decision offline. The hard
gate commits only when no blocker fires; the thresholded gates additionally require
$P(\mathrm{correct})\ge\tau$. We choose $\tau$ on a deterministic dev split and apply it
once to the held-out test split. Table~\ref{tab:c3r_threshold_selection} shows the two
pre-specified choices: a utility-selected threshold and a threshold matched to C2's dev
coverage. A smaller C3 pilot and prompt-template sanity checks are reported in
the appendix.

\begin{table}[t]
  \centering
  \caption{\textbf{C3-R threshold selection.} Thresholds are selected on the dev split and
  then applied to the held-out test split. Lower loss is better; the default abstention loss
  is $0.25$.}
  \label{tab:c3r_threshold_selection}
  \begin{adjustbox}{max width=0.82\linewidth}
    \begin{tabular}{lrrrrr}
\toprule
Rule & $\tau$ & Dev loss & Dev coverage & Test loss & Test coverage \\
\midrule
Dev utility & 0.750 & 0.244 & 0.117 & 0.237 & 0.134 \\
Coverage matched & 0.500 & 0.285 & 0.543 & 0.275 & 0.499 \\
\bottomrule
\end{tabular}

  \end{adjustbox}
\end{table}

On the held-out split (Table~\ref{tab:c3r_test_comparison}), C3-R suppresses
overconfident-wrong commitments more strongly than C2, but it does not make C3-R the best
overall policy under the default utility. The dev-utility threshold yields high selective
accuracy ($0.800$) and low OC-Wrong ($0.015$), but only by reducing coverage to $0.134$; its
paired loss difference versus C2 is small and inconclusive ($+0.008$, 95\% bootstrap
interval $[-0.013,0.028]$). The hard and coverage-matched variants retain more coverage
($0.499$) but have worse loss than C2. Thus C3-R is best read as a conservative safety
gate, not as a performance-improving policy: it sharply reduces unsafe commitments, but at
a real coverage cost.

\begin{table}[t]
  \centering
  \caption{\textbf{Held-out C3-R policy comparison.} C3-R is applied to the held-out split
  after dev-only threshold selection. C3-R reduces OC-Wrong most strongly, but the gain is
  paid for by lower coverage and shows no clear loss advantage over C2.}
  \label{tab:c3r_test_comparison}
  \begin{adjustbox}{max width=\linewidth}
    \begin{tabular}{lrrrrr}
\toprule
Policy & $n$ & Coverage & Sel. Acc. & OC-Wrong & Mean loss \\
\midrule
C0 & 335 & 1.000 & 0.633 & 0.209 & 0.261 \\
C2 & 335 & 0.561 & 0.691 & 0.054 & 0.228 \\
C3-R-hard & 335 & 0.499 & 0.497 & 0.015 & 0.275 \\
C3-R-dev-utility & 335 & 0.134 & 0.800 & 0.015 & 0.237 \\
C3-R-coverage-matched & 335 & 0.499 & 0.497 & 0.015 & 0.275 \\
\bottomrule
\end{tabular}

  \end{adjustbox}
\end{table}

\FloatBarrier

\section{Discussion}
\label{sec:discussion}

\paragraph{False fixed points and H-Risk.}
The experiments support a structural reading of false stability.
The H-Risk family is best read as a language for that structure rather than as a
score-level calibration metric.
Operationally, the false-fixed-point problem is that local stability need not imply
truth-tracking, and reduced sensitivity need not selectively repair wrong answers.
The LTI model provides the clean closed-loop setting in which stability, conditioning,
sensitivity, and innovation amplification can be separated before moving to the
output-level LLM audit proxy.
In the linear--Gaussian setting, the composite index $\Hrisk$ increases when the
closed-loop operator $\Phi$ approaches instability, when its conditioning worsens, and
when the innovation process exhibits large transient amplification; in our simulations
this coincides with regimes of miscalibration and poor closed-loop behavior.
In the LLM proxy study, the domain-wise proxy $H_{\mathrm{proxy}}(d)$ ranks domains where C2
gain over C0 is largest, with medical epidemiology and social statistics near the top and
technical standard and literature/media near the bottom.
Because the LLM proxy is domain-level and task-specific, its natural use is retrospective
audit prioritization: high $H_{\mathrm{proxy}}$ marks audited domains where policy-level
confidence variation and observed overconfident errors coincide.
Direct label-aware baselines such as C0 error and Brier risk are stronger pure ranking
scores in this small audit; $H_{\mathrm{proxy}}$ is useful because it exposes the two
ingredients we want to inspect---policy movement and overconfident-wrong mass. Direct
transformer-level estimation of $\Hrisk_{\mathrm{LLM}}$ remains a separate, operator-level
problem.

\subsection{Structural stability of high-confidence false judgments}

The sensitivity analysis from Section~\ref{subsec:internal-stability} clarifies the fragility question.
A natural hypothesis is that overconfident errors might coincide with internally ``unstable'' computation, in the sense
that the corresponding hidden states are more sensitive to small input perturbations than those leading to confidently
correct answers.
Across this experiment, that gap is not systematically supported at the measured scale
(Table~\ref{tab:sl_gap_cc_ocw}). This is the empirical basis for the stable-miscalibration
reading: high-confidence errors need not be the locally brittle cases. The result is
separate from output- or dialogue-level robustness tests, which ask whether answers persist
under contextual interference or challenge \cite{xu2026illusionsconfidence,
saadat2026certaintyrobustness}; here the question is whether hidden-state movement shows an
OCW-specific fragility gap.

The self-critical prompt lowers sensitivity overall, but that damping is not selective
repair. It changes the inference trajectory and the commitment policy, which is why the
abstention and utility analyses matter alongside the internal probe. Recent confidence
studies make the same caution useful: verbal confidence can be informative without being a
sufficient policy for risk-sensitive abstention
\cite{yoon2025reasoningmodelsbetterexpress,wang2026faithfulconfidence,xie2026knowwhenwrong}.

The C3 pilot and C3-R follow-up test the explicit feedback version: answer,
warrant critique, boundary check, contradiction check, and commitment or abstention.
The stricter C3-R gate is intentionally less subjective because its blockers and threshold
selection are fixed before looking at the held-out split.
It reduces overconfident-wrong commitments; under fair dev-split thresholding, however, the
policy buys safety by abstaining more and does not clearly beat the single-shot C2 prompt on
default-utility loss.
This makes Kantian feedback experimentally isolable as a commitment gate.

Appendix checks on TinyLlama-1B-Chat, Qwen2.5-3B-Instruct, denser $\sigma$ sweeps, and
curated semantic rewrites preserve the same qualitative picture: the CC--OCW gap remains
small relative to the absolute sensitivity scale.

\subsection{A high-SNR inertia hypothesis for Qwen2.5}

\begin{figure}[t]
  \centering
  \includegraphics[width=1.0\textwidth]{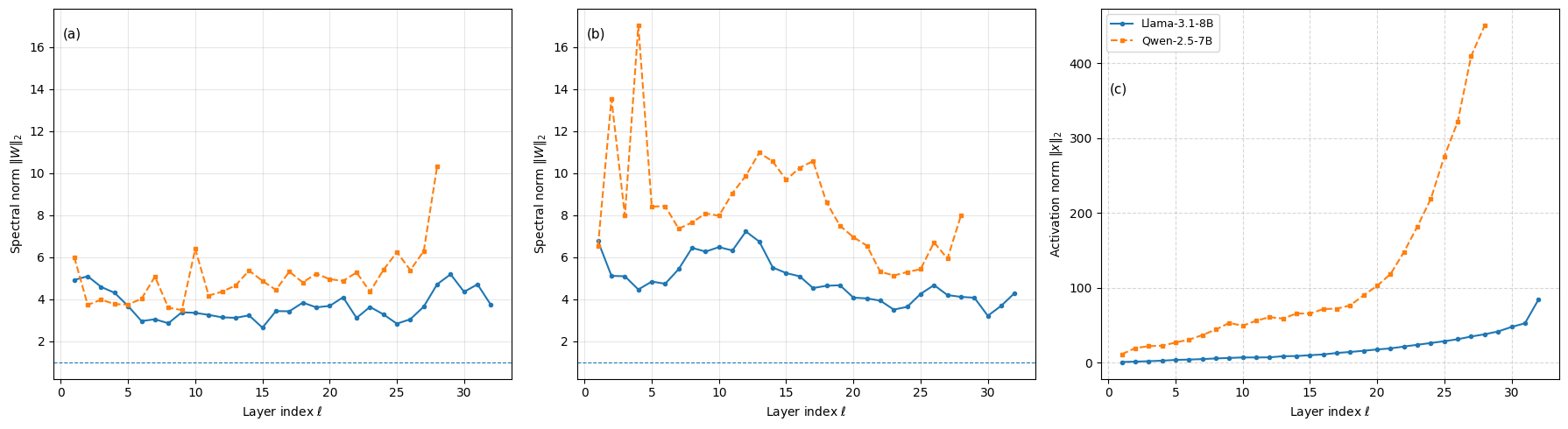}
  \caption{Spectral and activation profiles for Llama-3.1-8B and Qwen2.5-7B.
(a) Spectral norm $\lVert W\rVert_2$ of the attention output projections ($o_{\mathrm{proj}}$) across layers.
(b) Spectral norm of the MLP down-projections ($\mathrm{down}_{\mathrm{proj}}$).
(c) Layer-wise activation norm $\lVert x\rVert_2$ at the last token.
In all panels, solid lines (circles) denote Llama-3.1-8B and dashed lines (squares) denote Qwen2.5-7B.
Qwen2.5 exhibits consistently larger spectral norms and much larger activation norms,
indicating a high-SNR, low effective signal temperature regime in which relative perturbations are
strongly compressed despite large weights.}

  \label{fig:qwen_snr}
\end{figure}

Our sensitivity analysis shows that Qwen2.5 exhibits substantially lower local sensitivity
to input perturbations than Llama-3.1, even though our spectral measurements indicate that
Qwen's attention and MLP output matrices have \emph{larger} spectral norms
(Figure~\ref{fig:qwen_snr}(a,b)).
Large spectral norms are usually associated with signal amplification and potential instability, so this combination at
first looks paradoxical.

We resolve this tension by examining the magnitude of internal activations.
As shown in Figure~\ref{fig:qwen_snr}(c), Qwen2.5 maintains hidden states with much larger $\ell_2$ norms $\lVert
  x\rVert_2$ throughout the depth of the network.
In architectures with RMSNorm (as in Qwen2/Qwen2.5, which adopt RMSNorm in a pre-norm
Transformer design; \cite{qwen2_tech_report,qwen25_tech_report}), the normalization step
operates roughly as $x \mapsto x / \mathrm{RMS}(x)$ (up to a learned gain), so the effective
impact of a fixed-size perturbation $\epsilon$ on the normalized state scales like
$\epsilon / \mathrm{RMS}(x)$. Since $\mathrm{RMS}(x)=\lVert x\rVert_2/\sqrt{d}$ for a
$d$-dimensional state, our empirical $\lVert x\rVert_2$ profiles imply the same conclusion
up to a constant factor.
When $\lVert x\rVert_2$ is very large, the relative influence of $\epsilon$ is therefore strongly compressed.

This suggests a possible robustness channel: Qwen2.5 may damp the relative effect of a
fixed perturbation not through small weights, but through a high signal-to-noise regime in
which large-magnitude internal states endow the computation with strong ``inertia''.
In this sense Qwen2.5 behaves like a lower effective signal temperature
\footnote{This is an analogy for perturbation compression under normalization,
not the softmax temperature used in sampling or temperature scaling.}
system: typical small perturbations are diluted by normalization against very
large internal magnitudes, so the hidden-state trajectory is comparatively inert.

Importantly, this perturbation-compression channel can coexist with more discrete instability modes.
When internal activations and projection norms are large, attention logits can
become high-magnitude and the softmax can saturate, yielding near one-hot
attention patterns. Such saturation can make the computation appear locally
stable while creating a \emph{hard-switching} regime: rare perturbations that
change the top logit can produce abrupt downstream changes. One concrete path is
\emph{attention-head outliers}, where a small number of heads dominate
pre-softmax scores and produce large logit gaps (see the Qwen2.5 case study in
\cite{chen2025qwen25hugeflaw}). This ``quiet-then-flip'' behavior is distinct
from RMSNorm-based perturbation compression. It gives a second channel through
which a high-SNR model can look stable while retaining brittle decision
boundaries.

\enlargethispage{4pt}
\subsection{Representational compression as a possible mechanism}
\label{subsec:compression-mechanism}

The present sensitivity probe measures the magnitude of hidden-state movement under small perturbations.
A geometric diagnostic asks an additional question: whether the relevant alternatives occupy
a high- or low-dimensional region of representation space.
A stable high-confidence error may arise not only from attractor-like dynamics
or high-SNR inertia, but also from representational compression.
In this case, truth-relevant distinctions collapse into a low-effective-rank
semantic region. Nearby prompts and meaning-preserving rewrites move the state
within the same compressed basin, while the readout remains confident in the
wrong answer.

This compression-stabilized miscalibration hypothesis is compatible with work on
representation degeneration and anisotropy. Gao et al.\ report that NLG word
embeddings can concentrate in a narrow cone, limiting representation power
\cite{gao2019representationdegeneration}. Godey et al.\ frame Transformer
anisotropy as hidden representations becoming unexpectedly close in angular
distance \cite{godey2024anisotropy}. Recent representation-geometry work uses
effective rank and eigenspectrum decay to track representational collapse,
expansion, and compression across training
\cite{li2025representationgeometry,garrido2023rankme}.
Internal-state hallucination studies further show that truthfulness information can be
encoded in hidden representations even when external behavior is wrong, and that these cues
may travel through multiple question- and answer-anchored pathways
\cite{ji2024_internal_states_hallu_risk,orgad2025knowmore,luo2026twopathways}.
Our compression hypothesis is therefore not that truth information is absent; it is that the
local basin around some high-confidence errors may fail to expose truth-discriminative
directions to the commitment readout.

This hypothesis complements the high-SNR interpretation above.
High-SNR inertia makes a fixed perturbation relatively small compared with the state norm.
Representational compression predicts that even when the state moves, it may not
move along truth-discriminative directions.
Attractor-like stability then describes the local return to the same wrong decision region.
These are complementary ways a model can be stable without being correct.

One way to test the hypothesis is to measure the geometry of clean and perturbed
hidden states. For layer $\ell$ and group
$G\in\{\mathrm{CC},\mathrm{OCW}\}$, let $H_\ell(G)$ denote the hidden states and
let $C_\ell(G)$ be their empirical covariance.
The entropy effective rank is
\begin{equation}
\operatorname{erank}(C)
=
\exp\!\left(-\sum_j p_j\log p_j\right),
\qquad
p_j=\frac{\lambda_j}{\sum_k \lambda_k},
\end{equation}
where $\{\lambda_j\}$ are the covariance eigenvalues.
The participation ratio is
\begin{equation}
\operatorname{PR}(C)
=
\frac{\left(\sum_j \lambda_j\right)^2}{\sum_j \lambda_j^2},
\end{equation}
and a simple directional anisotropy score is
\begin{equation}
A(H)=
\left\|
\frac{1}{n}\sum_i \frac{h_i}{\|h_i\|}
\right\|.
\end{equation}
For perturbation or semantic-rewrite trajectories, define
\begin{equation}
\Delta h_{\ell,i}^{(r)}
=
h_{\ell,i}^{(r)} - h_{\ell,i}^{\mathrm{clean}},
\end{equation}
and compute the effective rank of the covariance of these local displacement vectors.
Compression-stabilized miscalibration predicts that OCW items may show
sensitivity norms comparable to CC items while having lower local tangent
effective rank, higher anisotropy, or weaker truth-separability.
Our current $\|\Delta h\|/\|\epsilon\|$ probe measures only movement magnitude,
not whether that movement spans truth-relevant directions.

\paragraph{Model-specific epistemic profiles.}
The combination of local sensitivity and high-SNR behavior suggests that current LLMs may
occupy distinct epistemic regimes.
DeepSeek-R1 displays relatively high local sensitivity and more modest activation norms,
corresponding to an internally reactive regime in which small perturbations can have
comparatively large effects on hidden states.
Qwen2.5, by contrast, combines low local sensitivity with large spectral and
activation norms, indicating a high-SNR, low effective signal temperature regime with
strong signal inertia: once its internal representations have settled, small
perturbations have little influence on the subsequent trajectory.
Llama-3.1 lies between these extremes, with intermediate sensitivity and activation norms,
and thus serves as a more balanced reference point in our experiments.
This kind of epistemic profiling may be useful when reasoning about which models are more
likely to exhibit reactive versus inertial patterns of error under different prompting
regimes.

\subsection{Future work}

The next step is to scale both the models and the diagnostics. Larger checkpoints,
different training regimes, and multimodal settings should test whether the same
stable-miscalibration pattern persists. On the metric side, operator-level
$\Hrisk_{\mathrm{LLM}}$ would require feasible Jacobian--vector or layer-wise
linearization tools. Representation-geometry diagnostics can then test whether stable
high-confidence errors occupy more compressed semantic regions than confidently correct
answers. Finally, controlled benchmarks should separate task hardness from
architecture-specific stability, while richer feedback gates should be compared against
single-shot prompt variants and external safety scaffolds.

\section{Conclusion}

This paper studies false stability in LLMs: the possibility that a model can be locally
robust, internally coherent, and confidently wrong.
The control model gives the clean abstraction: stable closed-loop behavior and correct
inference can diverge. The LLM probes give the empirical pressure point: in the tested
regimes, OCW items are not systematically more locally fragile than CC items.
Abstention-aware self-critique reduces overconfident wrong commitments by trading away coverage, and
$H_{\mathrm{proxy}}(d)$ remains a retrospective audit aid rather than a deployment
estimator.

The main lesson is therefore narrow but useful: robustness and truth-tracking should be
measured separately. The next steps are:
\begin{itemize}
  \item test whether the CC--OCW sensitivity pattern persists in larger and multimodal
  models;
  \item estimate operator-level LLM stability more directly with Jacobian or linearization
  tools;
  \item evaluate representation-geometry diagnostics such as effective rank, anisotropy,
  local tangent-rank, and truth-separability;
  \item compare explicit commitment gates against stronger non-Kantian abstention and
  critique baselines.
\end{itemize}

\section*{Limitations and Broader Impact}

\paragraph{Limitations.}
Our analysis has deliberately narrow scope. The linear--Gaussian model is a minimal
abstraction, not a mechanistic reduction of transformer computation; it omits nonlinear
dynamics, model misspecification, and multi-agent or social feedback. The LLM study uses a
small binary factual audit, a handful of open-weight models, and local hidden-state
sensitivity probes that are easiest to interpret in small-perturbation regimes. The domain
rankings and sensitivity pattern should therefore not be read as evidence for larger
datasets, multi-class tasks, long-form generation, or open-ended question answering. The
audit proxy $H_{\mathrm{proxy}}$ is label-aware and retrospective, not a deployment-time
uncertainty estimator, a superior labeled predictor, or a direct estimate of
$\Hrisk_{\mathrm{LLM}}$. The C3 and C3-R runs are post-hoc follow-ups on the frozen item set,
and the paper does not isolate a Kant-specific causal effect. Finally, representational
compression remains a proposed mechanism to test, not an empirical conclusion established
here.

\paragraph{Future directions.}
Future work should test larger and more diverse datasets, label-free deployment
approximations such as answer consistency, semantic entropy, retrieval disagreement, or
calibrated confidence proxies, and extensions beyond binary labels. Multi-class tasks can
use vector-valued proper scores; open-ended tasks will likely require semantic clustering,
judge-assisted correctness labels, or task-specific utility definitions.

\paragraph{Broader Impact.}
This work connects philosophy of cognition, control theory, and AI safety.
A stability-based view of hallucination may help practitioners look beyond scalar accuracy
and report calibration, uncertainty, and perturbation sensitivity. The risk is
over-interpretation: the formalism and model ``profiles'' are not safety guarantees or
normative rankings. The self-critical abstention policy gives modest, domain-dependent gains
and is not a sufficient safeguard in high-stakes settings. The Kantian translation is a
heuristic for scrutiny, not a source of authority.

\section{Reproducibility Checklist}
We provide code, data paths, and fixed seeds to reproduce all figures and tables in this manuscript.
\begin{itemize}
  \item \textbf{Repository:} public artifact branch:
  \href{https://github.com/ToppyMicroServices/202510_report_AI/tree/main}{%
  \texttt{ToppyMicroServices/202510\_report\_AI@main}}.
  \item \textbf{Environment:} Python \texttt{3.9.6}; dependencies in
  \path{requirements.txt}; build with \texttt{make v4}.
  \item \textbf{Artifact manifest:} \path{paper/latex_v3/ARTIFACTS.md} maps
  claims to frozen inputs, tables, figures, and scripts.
  \item \textbf{Data sources:} audit aggregates, the workshop CSV, and C3-R logs
  are listed in the artifact manifest.
  \item \textbf{Seeds \/ determinism:} the LTI simulation uses
  \texttt{CFG["seed"] = 2025} and shared noise sequences
  (\texttt{W\_SEQ}, \texttt{V\_SEQ}). The
  $\Delta\mathrm{SE}_{\mathrm{policy}}$ analysis is deterministic given the
  input CSV. The C3-R split seed is \texttt{c3r-dev-20260517}.
  \item \textbf{Figure scripts:} see the artifact manifest. Core frozen-input
  checks run with \texttt{make v4-artifacts}; optional LTI figures run with
  \texttt{python scripts/LTI.py}.
  \item \textbf{How to reproduce:} run \texttt{make v4-artifacts}, optionally
  run \texttt{python scripts/LTI.py}, then run \texttt{make v4}.
\end{itemize}
\paragraph{Computational note.}
The steady-state covariance $P$ is obtained by solving the discrete-time
Lyapunov equation $P=\Phi P\Phi^\top+\Sigma$ using the Bartels--Stewart
algorithm based on Schur decomposition \cite{golub2013}; existence and uniqueness of a 
positive-definite $P$ under $\rho(\Phi)<1$ follow from standard results in optimal filtering 
and Lyapunov stability theory \cite{anderson1979, zhou1996}.

\section*{Competing Interests}
\noindent\textbf{Author Note.} This work was conducted in a personal capacity,
outside the author's employment with another organization. ToppyMicroServices
O\"U is the author's independently owned, early-stage startup listed as a
correspondence affiliation; it is not the author's employer. No external funding
was received. The views expressed are solely those of the author, and any errors
are the author's alone. The author reports no other competing interests relevant
to this work.

\section*{Compliance Statement}
This personal research was conceived and completed outside the scope of the author's employment, 
using only personally owned hardware and personal cloud/accounts; no employer facilities, 
data, source code, or confidential information were used. To the author's knowledge, 
the work does not fall under any employer intellectual property assignment, work-for-hire, 
or similar clause, does not rely on proprietary materials of the employer, and does not use 
the employer's name, trademarks, or branding.

\bibliographystyle{unsrtnat}
\bibliography{refs}

\appendix
\section*{Appendix: Supplementary Robustness Checks}
\addcontentsline{toc}{section}{Appendix: Supplementary Robustness Checks}

\subsection*{Auxiliary feedback-loop and prompt checks}
\label{app:aux-feedback}

Before the full C3-R follow-up, we ran a small post-hoc C3 pilot on the first $100$ paired
items from the frozen audit set. C3 uses the released explicit loop: initial answer, warrant
critique, revision or abstention gate, and final confidence report. Table~\ref{tab:c3_feedback_pilot}
compares C3 with the existing C0--C2 logs on the same $100$ items. The result is a pilot:
C3 eliminates overconfident-wrong answers on this subset, but it does so by lowering
coverage, and its default-utility loss reduction is similar to C2 with a confidence interval
that includes zero.

\begin{table}[H]
  \centering
  \caption{\textbf{C3 explicit feedback-loop pilot on paired frozen-audit items.} C3 is a
  post-hoc $N=100$ run on items that already have C0--C2 logs. $\Delta SE$ is paired against
  C0 under the default abstention loss $0.25$ with 95\% bootstrap intervals.}
  \label{tab:c3_feedback_pilot}
  \begin{adjustbox}{max width=\linewidth}
    \begin{tabular}{lrrrrr}
\toprule
Policy & Coverage & Sel. Acc. & Answer Yield & OC-Wrong & $\Delta SE$ vs C0 \\
\midrule
C0 & 1.000 & 0.600 & 0.600 & 0.400 & 0.000 \\
C1 & 0.400 & 0.750 & 0.300 & 0.100 & -0.054 [-0.102, -0.008] \\
C2 & 0.500 & 0.680 & 0.340 & 0.160 & -0.039 [-0.086, 0.007] \\
C3 & 0.430 & 0.651 & 0.280 & 0.000 & -0.038 [-0.084, 0.008] \\
\bottomrule
\end{tabular}

  \end{adjustbox}
\end{table}

The repository also contains an earlier prompt-template run on general, logic, and reading
items. We treat it only as a sanity check because it is not the same frozen $11$-domain audit
used in the main text. In that run, generic critique-style templates slightly increase
uncertainty signalling without materially changing mean confidence, and a stronger
critique/noise template improves the available consistency score. Table~\ref{tab:legacy_prompt_ablation}
therefore supports a modest conclusion: prompt wording matters, but this legacy run does not
isolate a Kant-specific causal effect.

\begin{table}[H]
  \centering
  \caption{\textbf{Auxiliary legacy prompt-template ablation.} This earlier run uses
  different tasks and is not used as evidence for the main C0--C2 audit claims. Values are
  descriptive means over the available labeled fields.}
  \label{tab:legacy_prompt_ablation}
  \begin{adjustbox}{max width=0.86\linewidth}
    \begin{tabular}{lrrrr}
\toprule
Prompt family & $n$ & Conf. & Consistency & Uncertainty/refusal \\
\midrule
Baseline prompt & 300 & 0.875 & 0.989 & 0.025 / 0.165 \\
Generic critique template & 300 & 0.873 & 0.989 & 0.060 / 0.145 \\
Stronger critique/noise template & 300 & 0.874 & 1.000 & 0.060 / 0.160 \\
\bottomrule
\end{tabular}

  \end{adjustbox}
\end{table}

We also ran a small single-shot probe on the CritPt benchmark~\cite{zhu2025probing} (train
split, $N=70$) using \texttt{gpt-4.1-mini}. We evaluate C0 (forced answer; refusal
disallowed) versus C2 self-critical abstention. To make abstentions machine-detectable, the
first output line is constrained to be exactly \texttt{Answer.} or \texttt{Cannot judge.}

\begin{table}[H]
  \centering
  \small
  \caption{CritPt (train) auxiliary abstention probe with \texttt{gpt-4.1-mini}. The first
  line is constrained to \texttt{Answer.} or \texttt{Cannot judge.} (C0 forbids refusal).
  Brackets show 95\% item-level bootstrap confidence intervals. This probe measures
  refusal/hesitation behavior rather than benchmark accuracy.}
  \label{tab:critpt_abstention}
  \begin{tabular}{lccc}
    \hline
    Condition & Answer rate & Abstain rate & $N$ \\
    \hline
    C0 (forced Answer) & 1.000 & 0.000 & 70 \\
    C2 (self-critical abstention) & 0.329 [0.214, 0.443] & 0.671 [0.557, 0.786] & 70 \\
    \hline
  \end{tabular}
\end{table}

\subsection*{Noise-scale and semantic-rewrite checks}
\label{app:sensitivity-noise-scale}

The main internal-sensitivity probe uses Gaussian embedding noise with $\sigma=0.01$.
To check whether the absence of a clear CC--OCW fragility gap depends on this single
noise scale, we ran denser final-layer sweeps on the four-domain high-risk subset
(\texttt{medical\_epidemiology}, \texttt{social\_stats},
\texttt{history\_diplomacy}, \texttt{cultural\_industry}) for DeepSeek-R1 and
Qwen2.5-7B. The sweep uses
$\sigma\in\{0.0025,0.005,0.0075,0.01,0.015,0.025,0.0375,0.05\}$.
Across this range, the signed OCW--CC sensitivity gap remains small and non-monotone.

\begin{figure}[htbp]
  \centering
  \includegraphics[width=0.92\textwidth]{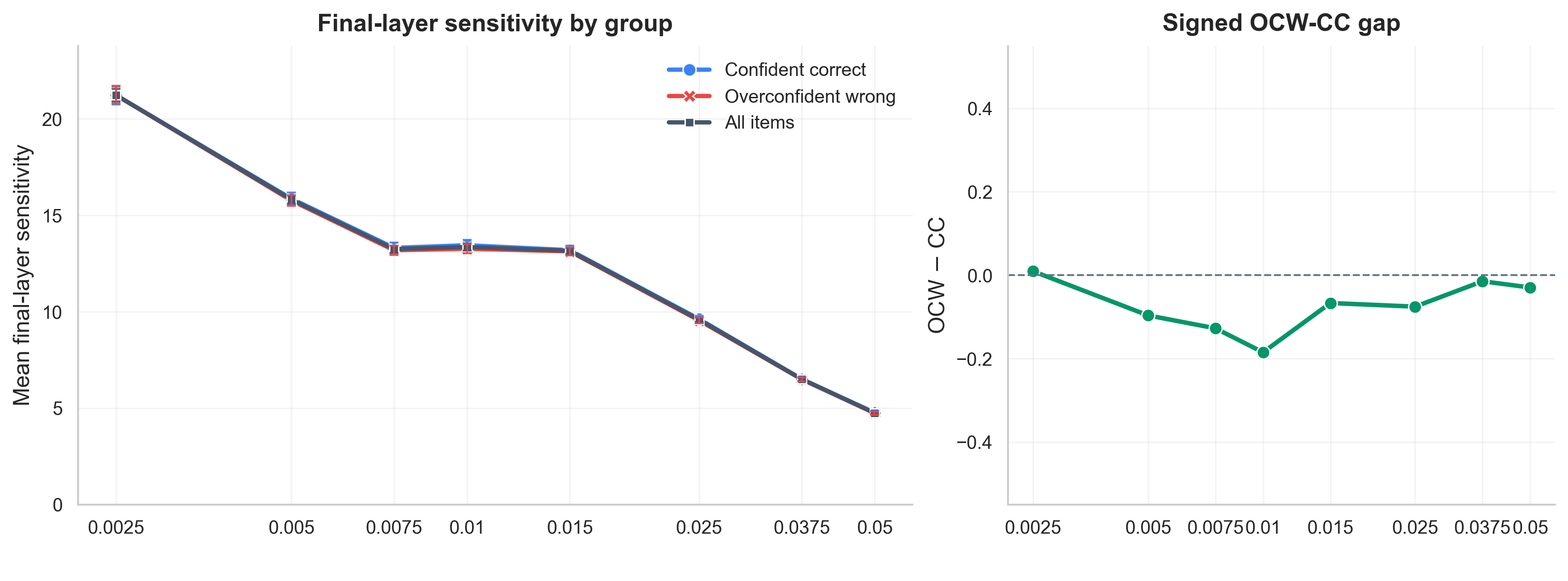}
  \caption{DeepSeek-R1 subset: final-layer sensitivity under larger Gaussian perturbations.
  Left: mean sensitivity for confidently correct (CC), overconfidently wrong (OCW), and
  pooled items as the embedding-noise scale increases. Right: signed OCW--CC gap.}
  \label{fig:sigma_sweep_deepseek}
\end{figure}

\begin{figure}[htbp]
  \centering
  \includegraphics[width=0.92\textwidth]{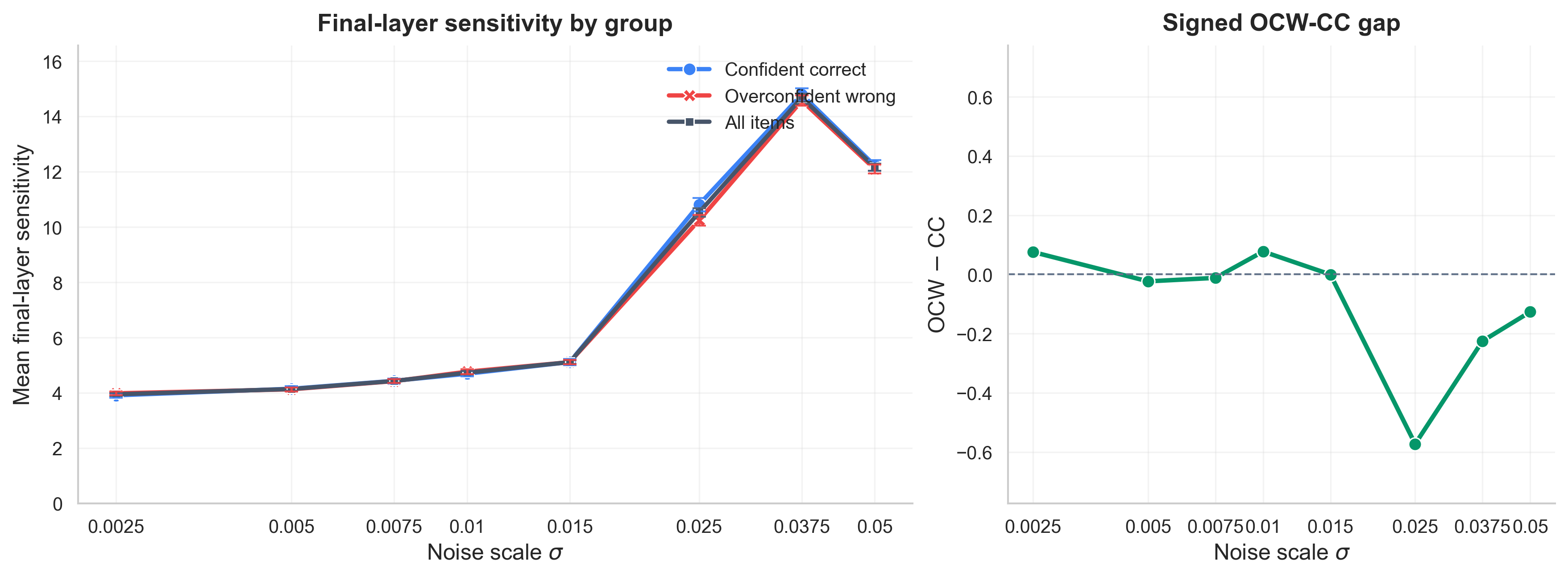}
  \caption{Qwen2.5-7B subset: final-layer sensitivity under larger Gaussian perturbations.
  The setup matches Figure~\ref{fig:sigma_sweep_deepseek}; the OCW--CC gap remains small
  across the tested noise scales.}
  \label{fig:sigma_sweep_qwen}
\end{figure}

We also tested curated meaning-preserving rewrites on the same two models. For DeepSeek-R1,
the final-layer relative semantic shift under the standard prompt was $0.285$ for CC items
and $0.307$ for OCW items (gap $+0.022$); under the self-critical prompt the means were
$0.156$ and $0.152$ (gap $-0.004$). For Qwen2.5-7B, the corresponding standard-prompt
values were $0.196$ and $0.221$ (gap $+0.025$), and the self-critical values were $0.122$
and $0.121$ (gap $-0.001$). These checks broaden the negative fragility result beyond
fixed-scale Gaussian noise.

\begin{figure}[htbp]
  \centering
  \begin{subfigure}[t]{0.49\textwidth}
    \centering
    \includegraphics[width=\linewidth]{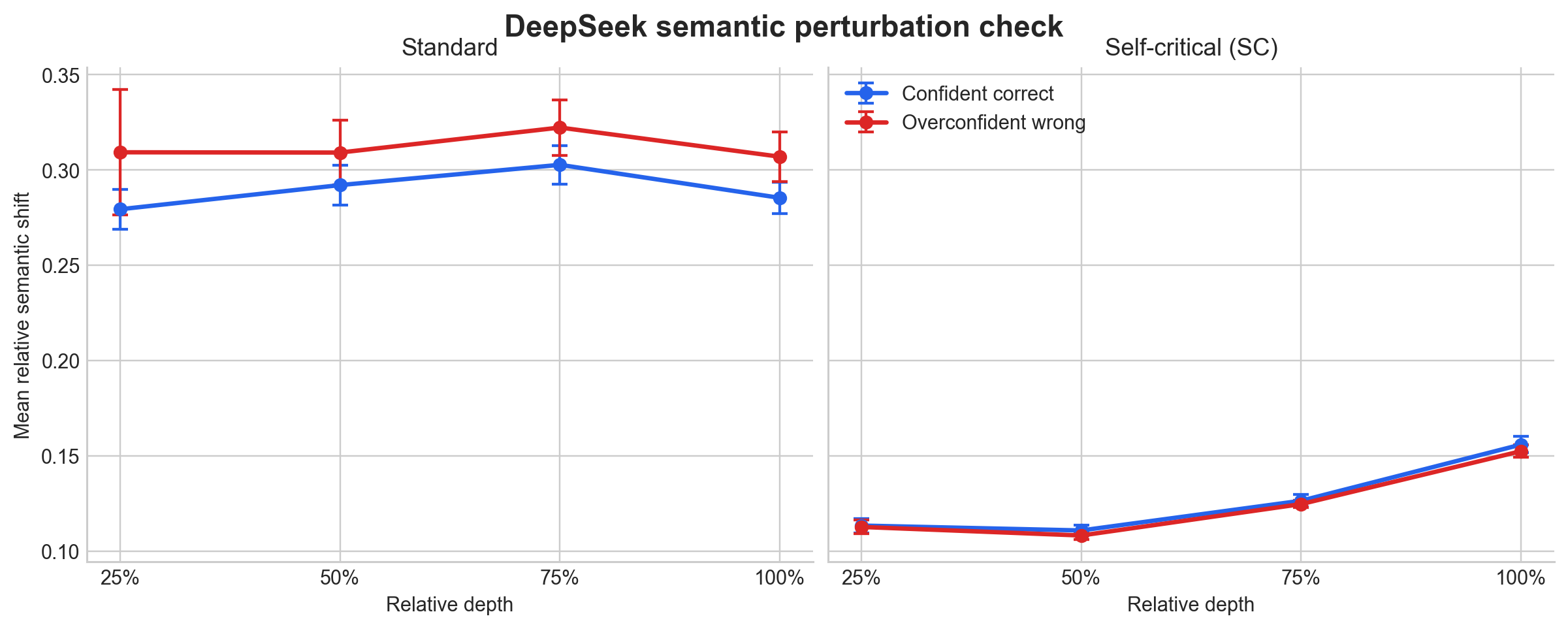}
    \caption{DeepSeek-R1}
    \label{fig:semantic_check_deepseek}
  \end{subfigure}
  \hfill
  \begin{subfigure}[t]{0.49\textwidth}
    \centering
    \includegraphics[width=\linewidth]{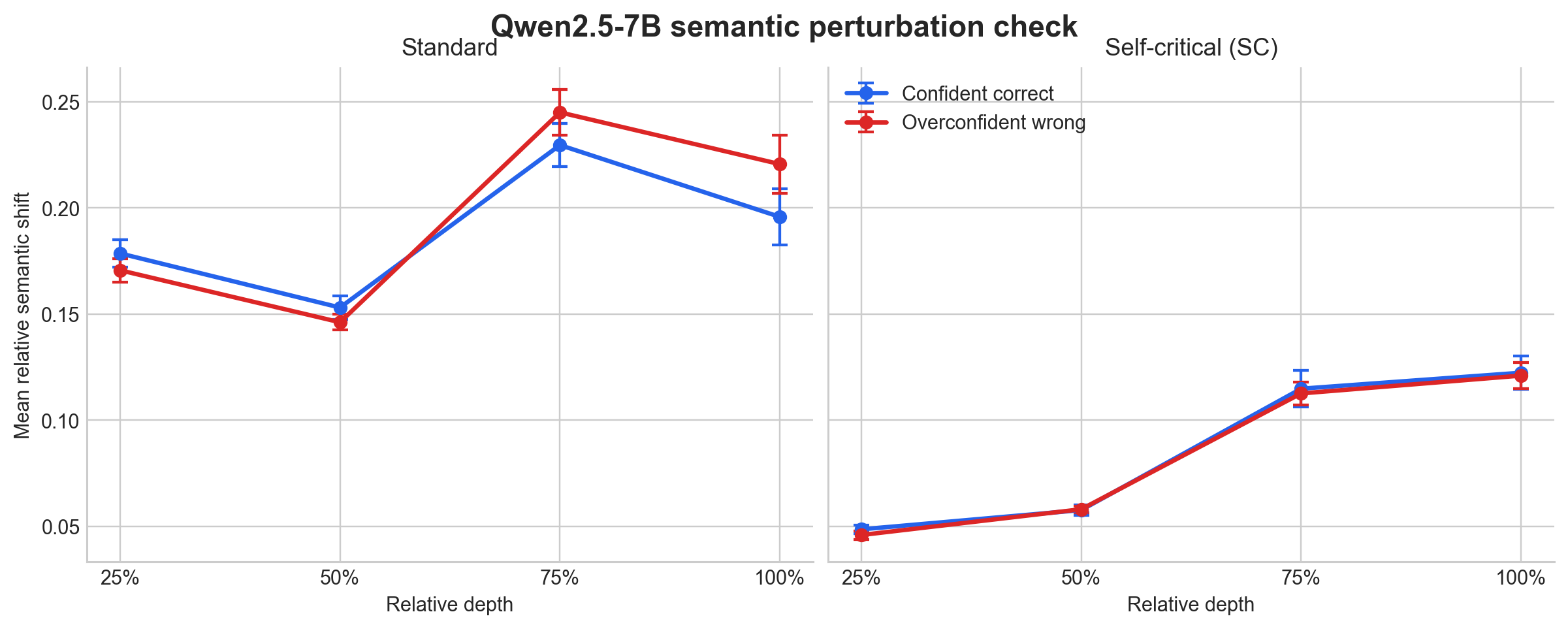}
    \caption{Qwen2.5-7B}
    \label{fig:semantic_check_qwen}
  \end{subfigure}
  \caption{Semantic perturbation checks. Panels show mean relative hidden-state shift under
  meaning-preserving rewrites for CC and OCW items across four representative depths.}
  \label{fig:semantic_checks}
\end{figure}

\subsection*{Smaller open-weight models}
\label{app:qwen_tiny_sensitivity}

To check whether the C0 sensitivity pattern is tied to the three 7--8B models in the main
text, we repeated the same final-layer analysis on TinyLlama-1B-Chat and
Qwen2.5-3B-Instruct, restricted to the same high-risk domains. In both cases the mean
sensitivity for overconfidently wrong items was within 2--3\% of that for confidently correct
items, and slightly lower. This supports the qualitative claim in
Section~\ref{subsec:internal-stability} without adding another full set of plots.

\end{document}